\documentclass{article}

\usepackage{PRIMEarxiv}
\usepackage[utf8]{inputenc} 
\usepackage[T1]{fontenc}    
\usepackage[colorlinks,linkcolor=red,anchorcolor=black,citecolor=green]{hyperref}       
\usepackage{url}            
\usepackage{booktabs}       
\usepackage{amsmath,amsfonts,amssymb}      
\usepackage{nicefrac}       
\usepackage{microtype}      
\usepackage{lipsum}
\usepackage{fancyhdr}       
\usepackage{graphicx}       
\usepackage{cite}
\usepackage{caption}
\usepackage{mathrsfs}
\usepackage{authblk} 
\usepackage{longtable}
\usepackage{cleveref}
\usepackage{comment}
\usepackage[linecolor=red,backgroundcolor=red!25,bordercolor=red,textwidth=21mm]{todonotes}
\crefname{figure}{}{}

\graphicspath{{media/}}     

\title{Reconstruction of Patient-Specific Confounders in AI-based Radiologic Image Interpretation using Generative Pretraining}

\author[1 $\dagger$]{Tianyu Han}
\author[2]{Laura Žigutytė}
\author[1]{Luisa Huck}
\author[1]{Marc Sebastian Huppertz}
\author[1]{Robert Siepmann}
\author[3]{Yossi Gandelsman}
\author[4,5]{Christian Blüthgen}
\author[1]{Firas Khader}
\author[1]{Christiane Kuhl}
\author[1]{Sven Nebelung}
\author[2,6,7, *]{Jakob Nikolas Kather}
\author[1, $\dagger$, *]{Daniel Truhn}
\date{}
\affil[1]{\small Department of Diagnostic and Interventional Radiology, University Hospital Aachen, Germany}
\affil[2]{\small Else Kroener Fresenius Center for Digital Health (EKFZ), Technical University Dresden, Dresden, Germany}
\affil[3]{\small Department of Electrical Engineering and Computer Science, UC Berkeley, CA, USA}
\affil[4]{\small Institute for Diagnostic and Interventional Radiology, University Hospital Zurich, University of Zurich, Zurich, Switzerland}
\affil[5]{\small Center for Artificial Intelligence in Medicine and Imaging (AIMI), Stanford University, Stanford, CA, USA}
\affil[6]{\small Department of Medicine I, University Hospital Dresden, Dresden, Germany}
\affil[7]{\small Medical Oncology, National Center for Tumor Diseases (NCT), University Hospital Heidelberg, Heidelberg, Germany}
\affil[$\dagger$]{\small Correspondence should be addressed to T.H. (than@ukaachen.de) and D.T. (dtruhn@ukaachen.de)}
\affil[*]{\small Shared Senior Authorship}

\begin{document}
\maketitle

\begin{abstract}
    \textbf{Detecting misleading patterns in automated diagnostic assistance systems, such as those powered by Artificial Intelligence, is critical to ensuring their reliability, particularly in healthcare. 
    Current techniques for evaluating deep learning models cannot visualize confounding factors at a diagnostic level. 
    Here, we propose a self-conditioned diffusion model termed DiffChest and train it on a dataset of 515,704 chest radiographs from 194,956 patients from multiple healthcare centers in the United States and Europe. 
    DiffChest explains classifications on a patient-specific level and visualizes the confounding factors that may mislead the model. 
    We found high inter-reader agreement when evaluating DiffChest's capability to identify treatment-related confounders, with Fleiss' Kappa values of 0.8 or higher across most imaging findings. 
    Confounders were accurately captured with 11.1\% to 100\% prevalence rates. 
    Furthermore, our pretraining process optimized the model to capture the most relevant information from the input radiographs. DiffChest achieved excellent diagnostic accuracy when diagnosing 11 chest conditions, such as pleural effusion and cardiac insufficiency, and at least sufficient diagnostic accuracy for the remaining conditions. 
    Our findings highlight the potential of pretraining based on diffusion models in medical image classification, specifically in providing insights into confounding factors and model robustness.
    }
\end{abstract}

\section*{Introduction}

Confounding factors in medicine are often present in input data and lead to spurious correlations that affect model predictions \cite{castro2020causality,glocker2021causality,zeng2022uncovering,mukherjee2022confounding}. 
For instance, radiological datasets that are used to train a neural network for the detection of pneumothorax, i.e., the pathologic collection of air in the pleural space between the inner chest wall and the outer lung parenchyma, often contain a high number of patients that have already been treated with a chest tube.
Thus the classifier model may learn to associate the presence of a chest tube with the pneumothorax, rather than the disease itself \cite{rueckel2020impact}. 
This is problematic, since the medical value of such a model would be the detection of patients suffering from pneumothorax before treatment begins, i.e., before the chest tube is inserted.
The presence of a chest tube is therefore a confounder for the presence of pneumothorax.

Recent studies on explainable AI offer ample further evidence of the presence of confounders in clinical image interpretation \cite{zhao2020training, de2023high, zech2018variable, degrave2021ai}. 
Metal tokens in radiographs, for example, are often used to indicate the left or right side of the body, and are a confounder for the presence of pulmonary disease such as COVID-19 \cite{zech2018variable,degrave2021ai}.
However, the discovery of these confounding factors and the visualization of possible confounders at a patient-specific level is challenging.

\begin{figure}[h!]
    \centering
    \scalebox{1}{
    \includegraphics[trim=35 90 220 10, clip, width=\textwidth]{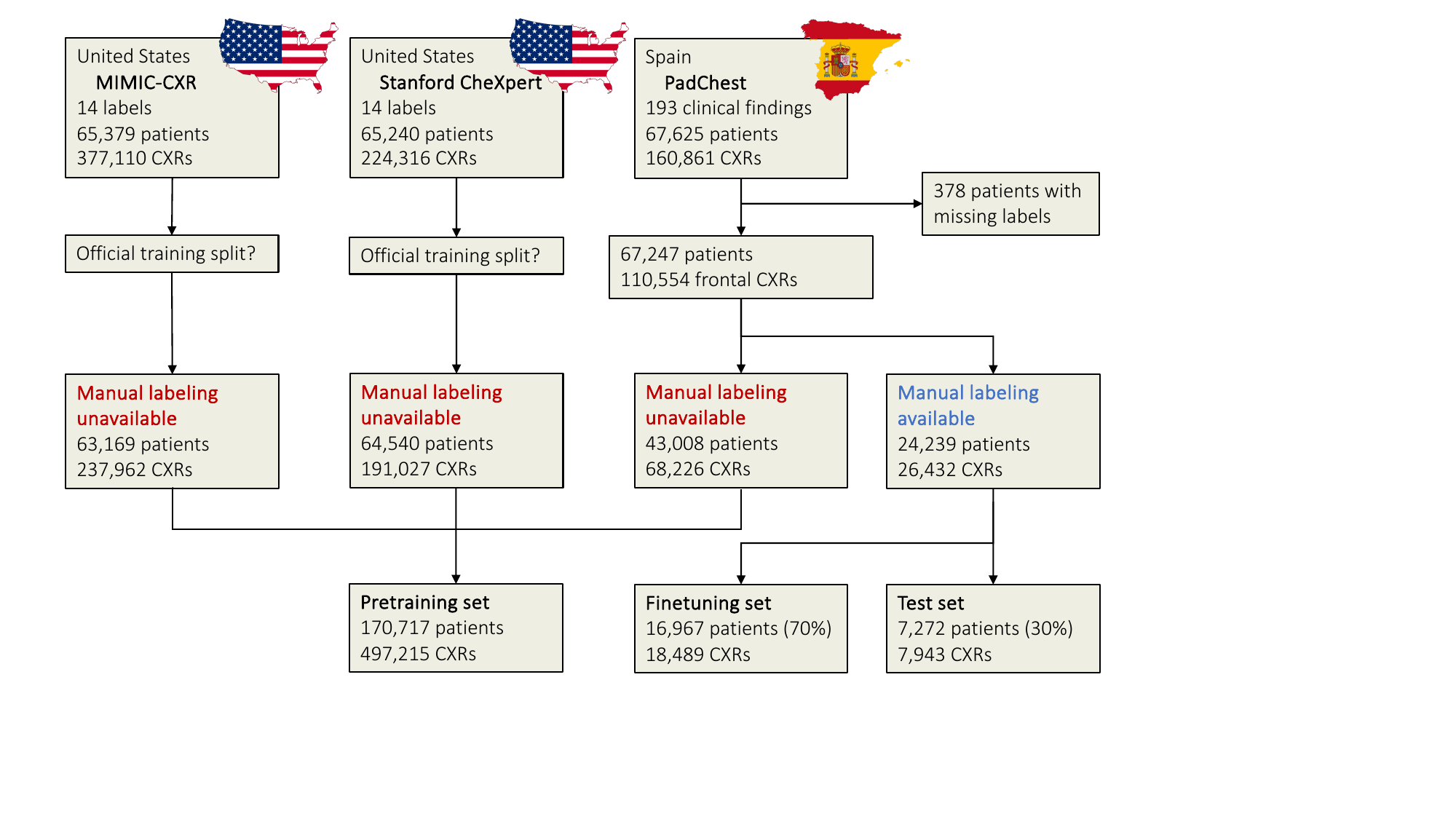}}
    \caption{\textbf{Study flow.}
    The DiffChest model was pretrained on a collection of 497,215 chest radiographs (CXRs) from California, Massachusetts, and Spain. 
    No labels were used to pretrain the model. 
    Subsequently, the model's classification head, consisting of a single logistic regression layer, underwent fine-tuning using a selected subset of 18,489 radiographs from the PadChest dataset, which had been manually annotated by clinicians.
    The model's performance was evaluated on a test cohort of 7,943 radiographs from the PadChest dataset, ensuring that these patients were not previously seen during both the pretraining and fine-tuning stages.}
    \label{fig:cohorts}
\end{figure}

One of the most promising developments in generative vision modeling are diffusion models \cite{ho2020denoising, ho2022imagen, li2022diffusion,khader2023denoising}.
These models offer a unique approach in which the original data is perturbed towards a tractable posterior distribution, often modeled as a Gaussian distribution. 
Subsequently, a neural network is employed to generate samples from this posterior distribution \cite{ho2020denoising}.
In our work, we propose a self-supervised pretraining framework that leverages diffusion models to generate patient-specific explanations for the classification of pulmonary pathologies.
After pretraining, diffusion models facilitate the generation of image representations that are not only rich in semantic information but also highly useful for disease classification.
We call our model DiffChest and train it on 515,704 chest radiographs (Figure \ref{fig:cohorts}).
The resulting model is both capable of synthesizing high-quality chest radiographs and exhibits a convolutional feature extractor for the classification of pulmonary pathologies.
The combination of the generative abilities with the classification capabilities of the model allows for generating patient-specific explanations with respect to 193 conditions and associated imaging findings. 
DiffChest further enables us to identify a wide range of confounders in the training data and to evaluate model biases and failures.
We underly our model with a theoretical foundation in which we prove that optimizing the diffusion objective is equivalent to directly maximizing the mutual information between the input and output of the feature extractor.
Our concept presents an approach where generative pretraining can effectively leverage vast amounts of unlabeled or noisy labeled data, enabling the development of efficient and interpretable model-assisted diagnosis in clinical settings.

\section*{Results}

\subsection*{Pulmonary Disease Detection}
\begin{figure}[h!]
    \centering
    \includegraphics[trim=0 5 20 0, clip, width=\textwidth]{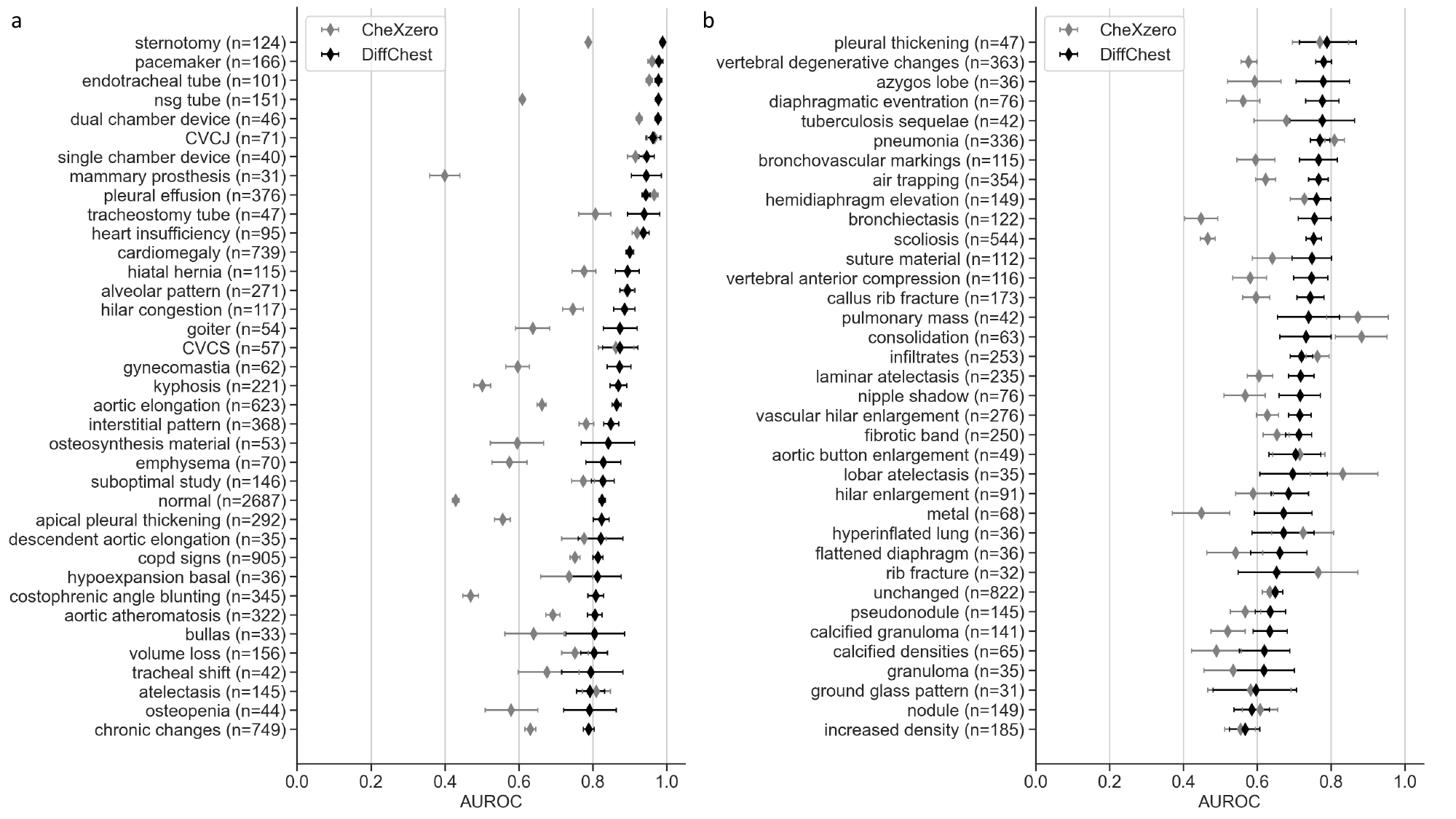}
    \caption{\textbf{AI-based detection of imaging findings by DiffChest and the CheXzero foundation model.}
    The mean AUC and 95\% CI is shown for each imaging finding with more than 30 entries in the PadChest testing cohort. 
    Findings are sorted by the mean AUC of DiffChest.
    Top 36 imaging findings are shown in panel \textbf{a}, and the rest imaging findings are shown in panel \textbf{b}.
    Among the total 73 evaluated imaging findings, DiffChest achieves a mean AUC of greater than 0.900 for 11 findings and greater than 0.700 for 59 findings.
    $n$ in this plot refers to the number of positive examinations in the PadChest test set.
    DiffChest outperforms CheXzero in 60 out of the 73 imaging findings.
    CVCJ$=$central venous catheter via jugular vein;
    CVCS$=$central venous catheter via subclavian vein;
    nsg tube$=$nasogastric tube;
    copd$=$chronic obstructive pulmonary disease.
    }
    \label{fig:vschexzero}
\end{figure}

We collected a testing cohort of 7,943 radiographs from 7,272 patients that were labeled by expert-clinicians, see Figure  \ref{fig:cohorts}.
We then evaluated DiffChest on this cohort for the task of finding the correct radiological diagnosis and compared its performance to the state-of-the-art foundation model \cite{moor2023foundation}, CheXzero, which previously demonstrated high performance on the PadChest dataset \cite{tiu2022expert}. 
CheXzero adopts contrastive learning, utilizing a dataset of more than 370,000 chest radiographs and radiology report pairs to learn a unified embedding space, enabling multi-label classification.
Figure \ref{fig:vschexzero} and Table \ref{tab:p_ci} illustrate the results of the comparison in terms of Area Under the Curve (AUC) and 95\% Confidence Intervals (CI) for 73 imaging findings.
To ensure a fair comparison, we only included findings with more than 30 entries in the PadChest testing cohort ($N=7,943$).
DiffChest outperforms CheXzero in 60 out of the 73 imaging findings, showcasing its superior performance. 
The average AUC improvement achieved by DiffChest over CheXzero is 0.110, with mean AUC values of 0.791 and 0.681, respectively.
DiffChest achieves an AUC of at least 0.900 for 11 imaging findings and at least 0.700 for 59 imaging findings out of the total 73 imaging findings evaluated, further underlining its potential applicability in clinical scenarios.

\subsection*{Generating Patient-Specific Explanations}

\begin{figure}[h!]
    \centering
    \includegraphics[trim=10 0 10 0, clip, width=\textwidth]{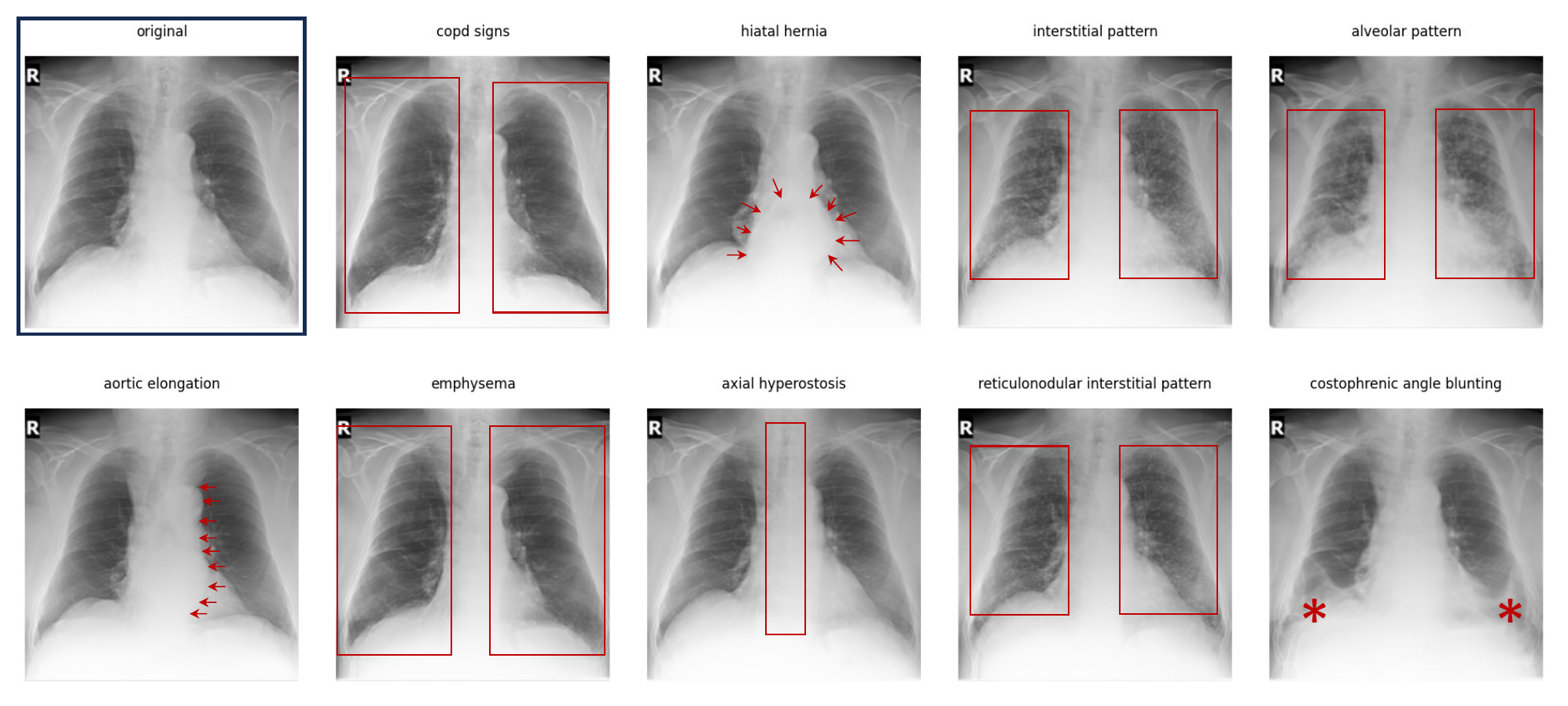}
    \caption{\textbf{Visual explanations of patient CXRs.}
    The original radiograph is highlighted by a black frame, which was acquired from a male patient born in 1928. 
    It was labeled as Normal in the PadChest dataset.
    To demonstrate that DiffChest incorporates the image characteristics of a multitude of pathologies, we synthesize nine different pathologies, namely chronic obstructive pulmonary disease signs, hiatal hernia, interstitial pneumonia, alveolar pneumonia, aortic elongation, emphysema, axial hyperostosis, reticulonodular interstitial pneumonia, and costophrenic angle blunting for this patient. 
    The generated explanations exhibit relevant pathological signatures, which are highlighted with arrows, stars, and frames by a board-certified radiologist with 12 years of experience.
    }
    \label{fig:counterfactuals}
\end{figure}

AI in medicine must be explainable to be integrated into clinical practice \cite{kundu2021ai}. 
While approaches like feature attribution \cite{sundararajan2017axiomatic}, class activation mapping (CAM) \cite{zhou2016learning}, and Grad-CAM \cite{selvaraju2017grad} provide insights by highlighting regions or pixels in an image that influence the classification, they fall short of explaining the specific attributes within those regions that are relevant to the classification. 
They do not clarify whether the classification of medical images is influenced by factors such as the radiographic density of an organ or its shape.
Here, we provide a visual explanation for DiffChest's classification decisions by starting from the representation of the original image in the semantic latent space $\mathbf{z}$ and then modifying this representation such that the image synthesized by the diffusion model has a higher probability of being classified as exhibiting the pathology (see Extended Data Figure \ref{fig:modelarch}c and Figure \ref{fig:counterfactuals}).
In other words, we produce a synthetic image that represents the same patient and that visualizes how the radiograph of that patient would look to the neural network if it had that pathology.

In detail, in the reverse diffusion kernel $p_\theta(\mathbf{x}_{t-1} | \mathbf{x}_t, \mathbf{z}^\ast)$, we add adversarial noise to the last logistic regression layer for the encoded feature $\mathbf{z}$:
\begin{equation}
\mathbf{z}^\ast := \mathbf{z} - \epsilon \cdot \nabla_{\mathbf{z}} \mathcal{L}(h(\mathbf{z}; \theta_{lr}), y^t),
\label{equ:adv}
\end{equation}
where $\mathcal{L}$ is the cross-entropy loss, $h(\mathbf{z}; \theta_{lr})$ is the logistic regression layer, and $y^t$ is the target label.
Unlike adversarial examples \cite{han2021advancing}, which are generated by perturbing the input image, our approach generates counterfactual examples by perturbing the latent code $\mathbf{z}$, which is then used to generate a new image containing the desired clinical attributes \cite{han2022image}. 
In Figure \ref{fig:counterfactuals} and Figure \cref{fig:more_counterfactuals_24,fig:more_counterfactuals_48,fig:more_counterfactuals_72}, our explanations answer the question: "Had the original $x$ been $x^t$, then the classifier prediction would have been $y^t$ instead of the original class $y$," where $x^t$ is the target class explanation and $y^t$ is the target label.

This paradigm is important for the introduction of AI models in medicine as it allows users to understand the model's decision-making by getting a direct visual feedback on which image characteristics lead to the model classifying a radiograph as exhibiting a certain pathology.
Examples visualized in Figure \ref{fig:counterfactuals} comprise overinflated lungs for patients suffering from chronic obstructive pulmonary disease, or characteristic changes of the lung parenchyma for patients suffering from interstitial pneumonic infiltration.
These visualizations might also allow the clinical expert to make more nuanced diagnoses by emphasizing different pneumonic patterns (compare interstitial pattern, alveolar pattern, and reticulonodular interstitial pattern) that would not have been possible with traditional methods of AI explainability such as CAM.

\subsection*{Confounders in the Training Data can be Identified}

\begin{figure}[h!]
    \centering
    \includegraphics[trim=45 45 290 0, clip, width=\textwidth]{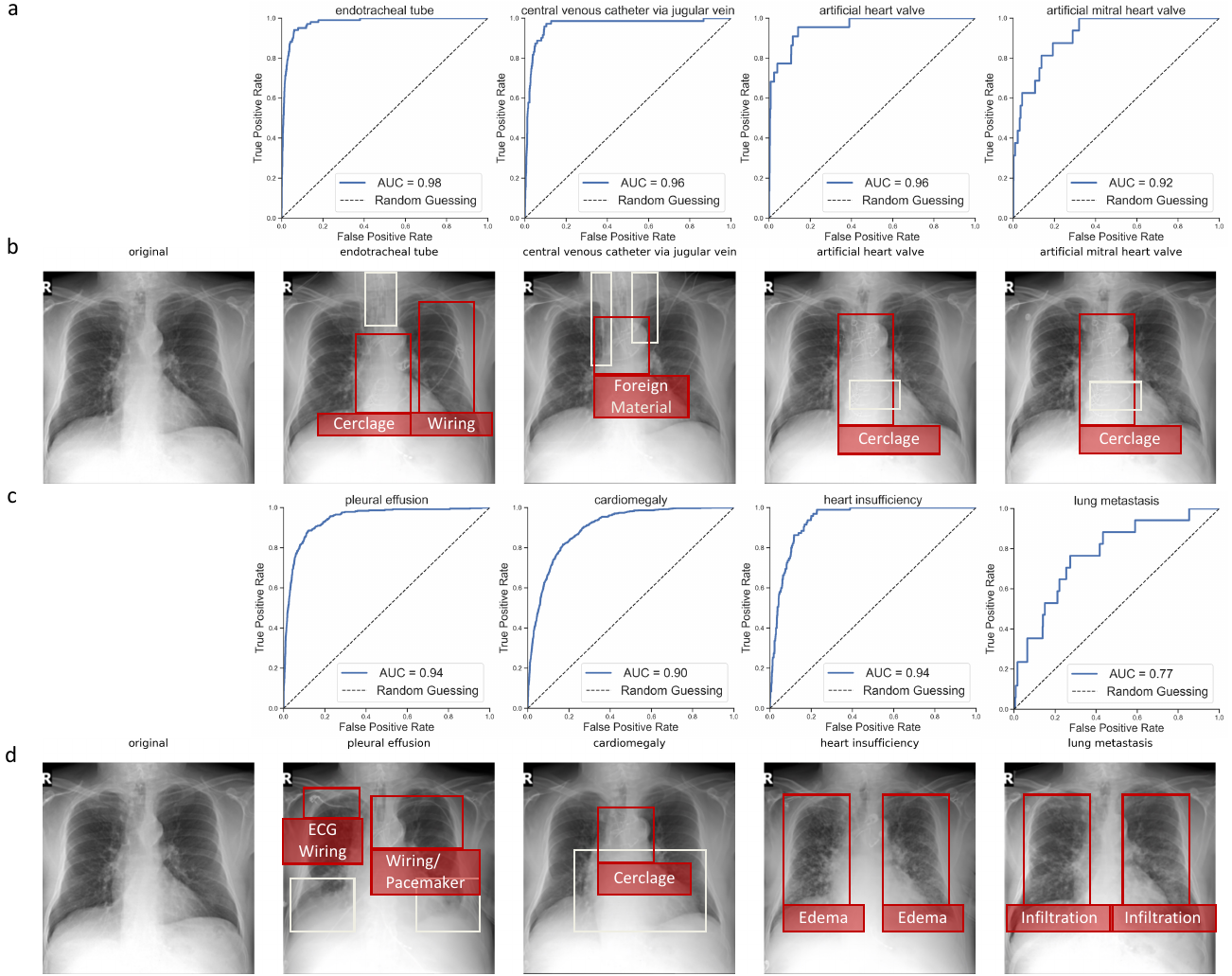}
    \caption{\textbf{Confounders as synthesized by DiffChest.}
    DiffChest achieves strong performance when acting as a classification model and detecting foreign materials (\textbf{a}) and the presence of pathologies as exemplified by a high AUC (\textbf{c}). 
    However, the model may rely on shortcut signals, i.e., confounders, to achieve this performance.
    We, therefore, employed DiffChest to synthesize radiographs that exhibit a specific pathology based on a non-pathological radiograph (\textbf{b} and \textbf{d}).
    The specific (target) pathology is highlighted by a white frame, while confounders are highlighted by a red frame.
    Typical confounding signals contained sternal wire cerclages, drainage tubes, device wirings, and catheters.
    Importantly, DiffChest also created synthetic radiographs that depict co-morbidities as shown in (\textbf{d}).
    }
    \label{fig:confounders}
\end{figure}

AI models are able to exploit subtle image characteristics for their diagnosis.
While this may be helpful to achieve super-human performance, it comes with a problem: When the AI model is trained to differentiate between two groups (e.g., patients with pneumonia and without pneumonia) it may focus on differences in the group that are not directly related to the pathology, but rather are present as spurious correlations.
For instance, patients suffering from pneumonia might more often have a catheter inserted, which is not directly related to pneumonia but rather to the treatment of pneumonia.
Therefore, an AI model which is trained to detect pneumonia may in fact learn to detect the presence of a catheter, which is a confounder \cite{zhao2020training, de2023high, zech2018variable, degrave2021ai}.
DiffChest's visual explanations allow us to identify such confounders by testing if these spurious correlations are present in these synthetic images.

To demonstrate this, we performed the following reader experiments:
We let DiffChest generate synthetic radiographs that exhibited a specific pathology based on a non-pathological radiograph.
Subsequently, a board-certified radiologist with twelve years of experience checked for the presence of confounders in these images.
Confounders were classified in either being treatment-related (such as the presence of a catheter for patients suffering from pneumonia, Figure \ref{fig:confounders}b) or as being related to the pathology (such as the presence of a lung congestion for patients suffering from cardiomegaly, Figure \ref{fig:confounders}d).
Figure \ref{fig:confounders} shows exemplary results for the generated synthetic images visualizing the target pathology and along with the confounders.

\begin{figure}[h!]
    \centering
    \scalebox{0.9}{
    \includegraphics[trim=0 0 0 0, clip, width=\textwidth]{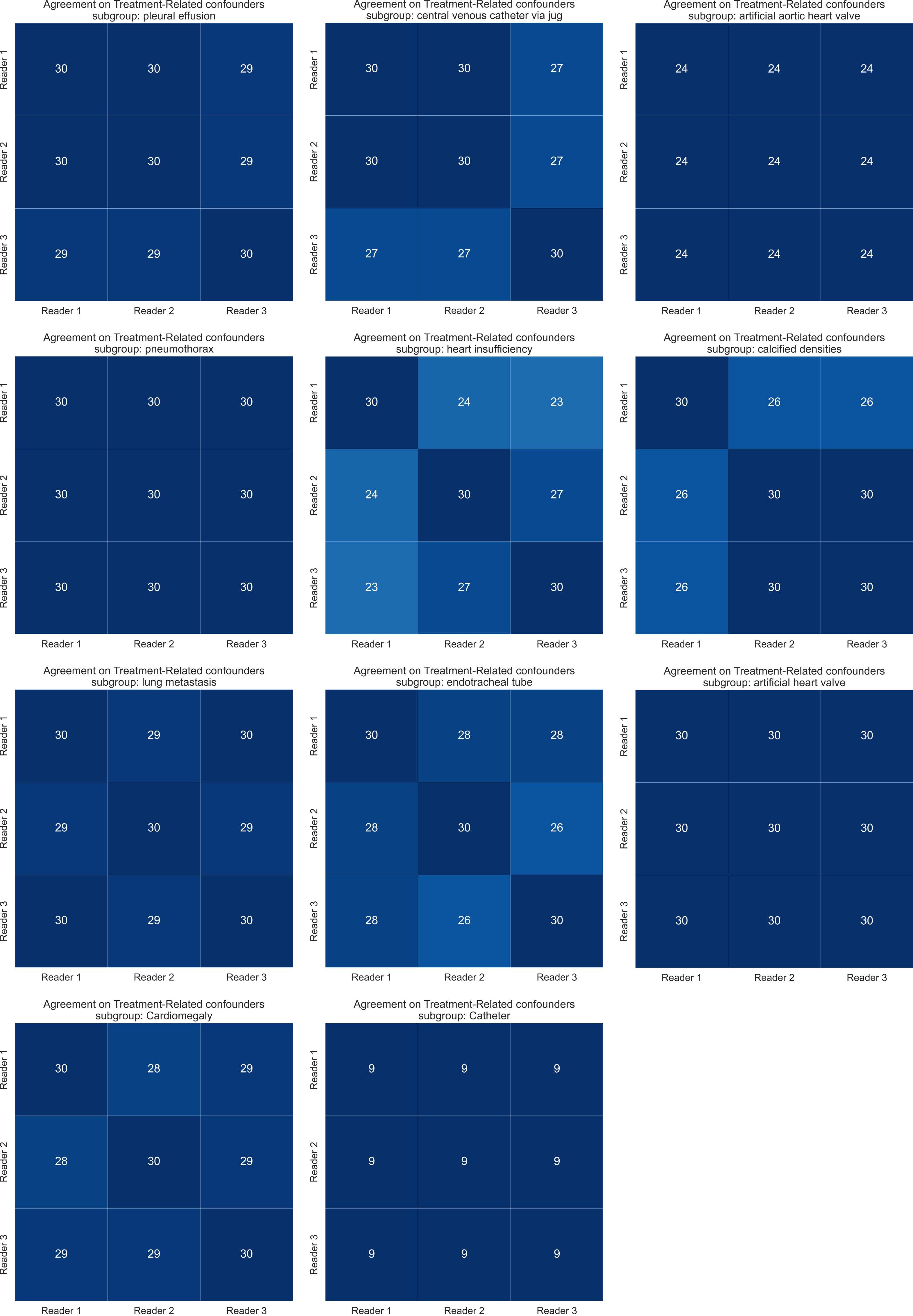}}
    \captionof{table}{\textbf{Inter-reader agreement on treatment-related confounders.}
    Each cell represents the number of radiographs that are rated as having the corresponding confounder by both radiologists.
    }
    \label{fig:reader_treat}
\end{figure}

\begin{figure}[h!]
    \centering
    \scalebox{0.95}{
    \includegraphics[trim=0 0 0 0, clip, width=\textwidth]{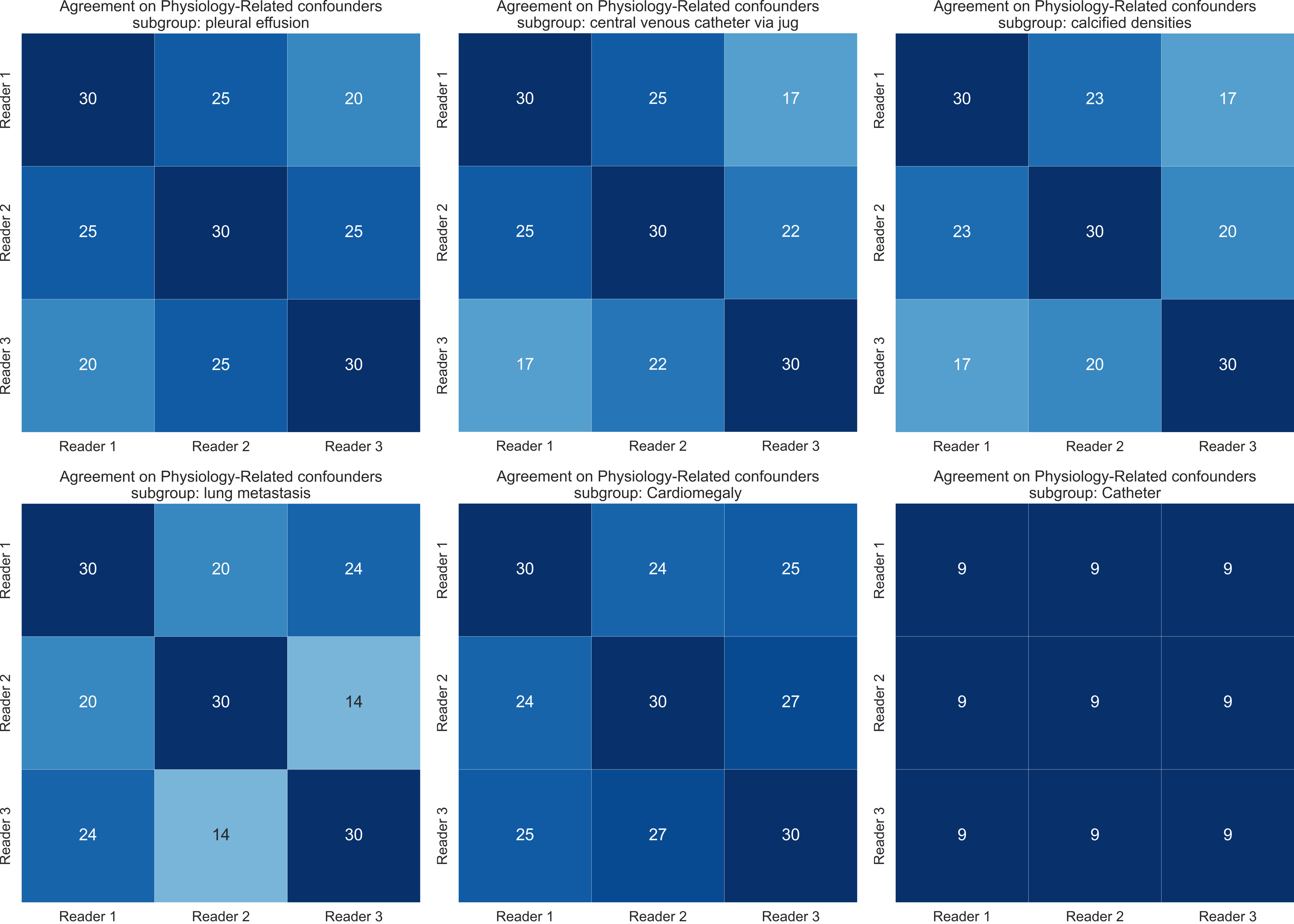}}
    \captionof{table}{\textbf{Inter-reader agreement on physiology-related confounders.}
    Each cell represents the number of radiographs that are rated as having the corresponding confounder by both radiologists.
    }
    \label{fig:reader_physio}
\end{figure}

To verify if our model can accurately pick up a confounding bias in the original data, we performed the second reader study in which we asked three radiologists to annotate the presence of confounders in the original data, see Extended Data Figure \ref{fig:readerstudy} for a detailed breakdown of the study design.
The results of this experiment are given in Tables \ref{tab:treat} and \ref{tab:physio} that give the number of real radiographs for which at least one treatment-related confounder (Table \ref{tab:treat}) or at least one physiology-related confounder (Table \ref{tab:physio}) was present.
We illustrated the discordance between the radiologists when identifying treatment-related and physiology-related confounders in Extended Data Figure \ref{fig:reader_treat} and \ref{fig:reader_physio}.
We found that the confounders that had been identified in the synthetic images were also present in a large proportion of the real radiographs, thereby proving that DiffChest is able to identify confounders in the data.


\begin{table}
    \centering
	\captionof{table}{\textbf{Presence of treatment-related confounders in radiographs}}
    \begin{tabular}{lrrrr}
        \toprule
        "Diagnosis" in Radiograph &  Radiographs (n) &  Confounder present &  Ratio (\%) &  Fleiss' Kappa \\
        \midrule
        artificial heart valve &         30 &                         30 &    100.0 &         1.000 \\
        central venous catheter via jugular vein &         30 &                         29 &     96.7 &         0.900 \\
        endotracheal tube &         30 &                         28 &     94.4 &         0.867 \\
        heart insufficiency &         30 &                         24 &     82.2 &         0.731 \\
        artificial aortic heart valve &         24 &                         22 &     91.7 &         1.000 \\
        pleural effusion &         30 &                         12 &     42.2 &         0.967 \\
        lung metastasis &         30 &                          6 &     21.1 &         0.967 \\
        pneumothorax &         30 &                          5 &     16.7 &         1.000 \\
        cardiomegaly &         30 &                          4 &     13.3 &         0.932 \\
        calcified densities &         30 &                          3 &     11.1 &         0.861 \\
        catheter &          9 &                          1 &     11.1 &         1.000 \\
        \bottomrule
    \end{tabular}
    \caption*{\small 
    In total, 303 radiographs were analyzed by three radiologists for the presence of a subset of confounders (Extended Data Figure \ref{fig:daniel_con}). The subset of confounders that these radiographs were tested for had previously been identified with the help of DiffChest.
    Fleiss' kappa was computed to quantify inter-reader agreement.
	}
    \label{tab:treat}
    
\end{table}

\begin{table}
    \centering
	\captionof{table}{\textbf{Presence of physiology-related confounders in radiographs}}
    \begin{tabular}{lrrrr}
    \toprule
    "Diagnosis" in Radiograph &  Radiographs (n) &  Confounder present &  Ratio (\%) &  Fleiss' Kappa \\
    \midrule
    central venous catheter via jugular vein &         30 &                         20 &     66.7 &         0.551 \\
    pleural effusion &         30 &                         20 &     66.7 &         0.660 \\
    cardiomegaly &         30 &                         17 &     58.9 &         0.766 \\
    calcified densities &         30 &                         14 &     46.7 &         0.466 \\
    lung metastasis &         30 &                         11 &     37.8 &         0.358 \\
    catheter &          9 &                          2 &     22.2 &         1.000 \\
    \bottomrule
    \end{tabular}
    \caption*{\small 
    In total, 159 radiographs were analyzed by three radiologists for the presence of a subset of confounders (Extended Data Figure \ref{fig:daniel_con}). The subset of confounders that these radiographs were tested for had previously been identified with the help of DiffChest.
    Fleiss' kappa was computed to quantify inter-reader agreement.
    }
    \label{tab:physio}
\end{table}

\subsection*{Mutual Information Maximization for Free}
Diffusion models belong to a class of latent variable models expressed as $p_{\theta}(\mathbf{x}_0) = \int p_{\theta}\left(\mathbf{x}_{0:T}\right) d\mathbf{x}_{1:T}$. 
In this formulation, $p_{\theta}(\mathbf{x}_i)$ refers to a neural network responsible for denoising a latent variable $\mathbf{x}_i$ at time $t=i$, as shown in Extended Data Figure \ref{fig:modelarch}a. 
The sequence of latent variables $\mathbf{x}_1$, $\mathbf{x}_2$, ..., $\mathbf{x}_T$ shares the same dimensionality as the observed data $\mathbf{x}_0$, sampled from the data distribution $p(\mathbf{x}_0)$ \cite{ho2020denoising}. 
However, such a high-dimensional latent space poses challenges in finding meaningful directions, i.e., directions which are identified by human experts as relevant to the disease. This property is, however, essential for representation control \cite{kim2022diffusionclip, kwon2022diffusion}. 
In DiffChest, we explicitly introduce an additional 512-dimensional latent space $p_\phi(\mathbf{z}|\mathbf{x}_0)$, which conditions the reverse diffusion process $p_\theta(\mathbf{x}_{t-1}|\mathbf{x}_{t}, \mathbf{z})$, to obtain a latent space in which we can identify meaningful directions. We will refer to this latent space as semantic latent space \cite{preechakul2022diffusion}.

Following the approach described in \cite{ho2020denoising} and \cite{preechakul2022diffusion}, DiffChest was trained using a simplified diffusion objective:
\begin{equation}
L_\text{simple} = \sum_{t=1}^{T} \mathbb{E}_{\mathbf{x}_0, \mathbf{\epsilon}_t} \left[ \left\| \mathbf{\epsilon}_\theta(\mathbf{x}_t, t, \mathbf{z}) - \mathbf{\epsilon}_t \right\|_2^2 \right].
\label{equ:loss}
\end{equation}
In Section \nameref{sec:ext_deri}, we prove that optimizing Equation \ref{equ:loss} is equivalent to maximizing the likelihood of a joint distribution between ${ \mathbf{x}_0, \mathbf{z} }$, i.e., $\log p_{\theta, \phi}(\mathbf{x}_0, \mathbf{z})$. 
This joint probability can be further decomposed into the product of the likelihood of the data $p_\theta(\mathbf{x}_0)$ and a second term $p_{\theta, \phi}(\mathbf{z}|\mathbf{x}_0)$, which is the posterior distribution of $\mathbf{z}$ given $\mathbf{x}_0$. 
Here, the second term $\log p_{\theta, \phi}(\mathbf{z}|\mathbf{x}_0)$ is also known as a lower bound of mutual information in Variational Information Maximization (VIM) \cite{barber2004algorithm, chen2016infogan}:
\begin{equation}
I(\mathbf{z}; \mathbf{x}_0) \geq \mathbb{E}_{(\mathbf{z}, \mathbf{x_0}) \sim p(\mathbf{z}, \mathbf{x_0})} \left[ \log p_{\theta, \phi}(\mathbf{z} | \mathbf{x}_0) \right] + H(\mathbf{z}),
\label{equ:mutualinfo}
\end{equation}
We denote $I(\mathbf{z}; \mathbf{x}_0)$ as the mutual information between $\mathbf{z}$ and $\mathbf{x}_0$, while $H(\mathbf{z})$ represents the entropy of $\mathbf{z}$. 
Therefore, we can interpret the training of DiffChest as maximizing both the data likelihood $\log p_\theta(\mathbf{x}_0)$ and the lower bound of the mutual information between the input image $\mathbf{x}_0$ and its encoded latent code $\mathbf{z}$. 
Once trained, DiffChest excels at generating high-quality samples while also preserving the semantic manipulation of the input data.

\section*{Discussion}


In this work, we used a diffusion pretraining method for medical image interpretation that allows to train classifiers without requiring large amounts of annotated data while also giving detailed insights into the model's decision-making process.
The model development involved large-scale pretraining on 497,215 unlabeled chest radiographs followed by a subsequent finetuning step on a small subset of annotated data.
The results show that, with limited finetuning, DiffChest outperforms CheXzero \cite{tiu2022expert}, an expert-level foundation model, in detecting 60 out of 73 pathologic imaging findings of the chest. 
Our method maintains high performance even when finetuning on only 3\% of the total finetuning data, demonstrating its data efficiency.

In addition, we demonstrated that the diffusion part of the model can be used in conjunction with the classification part to synthesize high-quality radiographs.
This capability can serve two purposes: First, it can explain the model's reasoning to its users and second it can be used to identify confounders in the training data.

This is important since medical practitioners need to have confidence in the rationale behind an AI model's predictions before they will consider delegating aspects of their workload to the system. 
The ability to view a synthesized image that highlights the features upon which the model focusses can facilitate a quick assessment by physicians to determine whether the AI model's diagnosis aligns with their own expert judgment.
Moreover, the issue of confounding variables poses a significant challenge in the development of AI models for medical applications, leading to potential model biases \cite{seyyed2021underdiagnosis}. 
The synthesization of radiographs by our model allows the direct identification of these confounders within the training dataset, thereby offering an avenue to enhance data quality and to build more robust machine learning models. 
We substantiated this claim through a reader study, wherein we demonstrate that the confounders identified by DiffChest were consistently present in a large subset of real radiographic images.

In contrast to currently prevailing contrastive learning approaches \cite{tiu2022expert,azizi2023robust}, which rely on the information noise contrastive estimation loss to maximize the mutual information between the input and the context vector \cite{oord2018representation,he2020momentum,chen2020simple,bachman2019learning}, we theoretically prove that optimizing the simplified diffusion objective while conditioning on the latent vector $\mathbf{z}$ is equivalent to maximizing both the data likelihood and the mutual information between the input and its encoded feature.
This theoretical insight lends support to the application of diffusion pretraining across the broader field of medical imaging.

However, our model's ability to manipulate attributes of real clinical images raises ethical concerns, similar to DeepFake \cite{nguyen2022deep}. 
This potential may in principle be exploited for the generation of fake medical data which may be leveraged for insurance fraud.
This is a problem since the detectability of manipulated samples is challenging as compared to standard adversarial approaches. 
A potential solution to this problem is the addition of another neural network for detection of fake images, as suggested by Preechakul et al. \cite{preechakul2022diffusion}.
Our work has limitations in that the resolution of the synthesized images is lower than what is normally used in clinical practice.
This may be solved by integrating progressive growing techniques as demonstrated by Ho et al. \cite{ho2022cascaded} and Karras et al. \cite{karras2017progressive}. 
However, implementing these techniques may require substantial hardware resources, which may not always be readily available in clinical settings. 
Exploring resource-efficient alternatives to improve generative capabilities while maintaining practical feasibility will therefore be an important future research direction.

In conclusion, this study demonstrates the potential of generative pretraining for developing interpretable models in medical image analysis. 
The model's ability to identify confounders in the training data and analyze model biases and failures is a significant step towards ensuring reliable and accountable AI applications in healthcare. 
Beyond chest radiography, our proof-of-concept study highlights the broader potential of self-supervised learning in conjunction with diffusion models, offering a pathway to  informed predictions for medical personnel.

\clearpage
\section*{Methods}
\subsection*{Ethics statement}
This study was carried out in accordance with the Declaration of Helsinki and local institutional review board approval was obtained (EK028/19).

\subsection*{Datasets}

Our method was trained on a diverse set of three publicly available datasets: the MIMIC-CXR dataset, the CheXpert dataset, and the PadChest dataset. 
The MIMIC-CXR dataset comprises an extensive collection of 377,110 frontal and lateral chest radiographs from 227,835 radiological studies \cite{johnson2019mimic}. 
To ensure the inclusion of unique patient information, we selected only the frontal radiographs, specifically the anteroposterior (AP) or posteroanterior (PA) views, for model pretraining, in cases where multiple radiographs were available for the same patient.
Within the MIMIC-CXR dataset, each radiograph has been meticulously labeled using an NLP labeling tool \cite{peng2018negbio} to indicate the presence of 14 different pathological conditions. 
These conditions include Atelectasis, Cardiomegaly, Consolidation, Edema, Enlarged Cardiomediastinum, Fracture, Lung Lesion, Lung Opacity, No Finding, Pleural Effusion, Pleural Other, Pneumonia, Pneumothorax, and Support Devices.

The CheXpert dataset, a large and publicly available dataset, contributed 224,316 chest radiographs obtained from 65,240 patients \cite{irvin2019chexpert}. 
Like the MIMIC-CXR dataset, the radiographs in the official training split of CheXpert have also been annotated using an NLP labeling tool \cite{irvin2019chexpert} to identify the presence of the same 14 pathological conditions as in the MIMIC-CXR dataset. 
In the official test split of CheXpert, radiological ground truth was manually annotated by five board-certified radiologists using majority vote \cite{irvin2019chexpert}. 
Manually annotated CheXpert radiographs were used for evaluating the data efficiency of our model, in section \nameref{sec:data_efficiency}.
The study flow of CheXpert is shown in Figure \ref{fig:chexpert_data}.

Another dataset used in our study is PadChest, a publicly available dataset comprising 160,861 chest radiographs from 67,625 patients at Hospital San Juan, Spain \cite{bustos2020padchest}. 
The radiographs in PadChest were extensively annotated, covering 193 different findings, including 174 distinct radiographic findings and 19 differential diagnoses \cite{bustos2020padchest}. 
Notably, 27\% of the images were manually annotated by trained physicians, while the remaining 73\% were automatically labeled using a recurrent neural network \cite{bustos2020padchest}.
We divided the PadChest dataset into two subsets based on patients: The first subset, comprising 43,008 patients, contained only automatically labeled images, while the second subset, consisting of 24,239 patients, contained only physician-labeled images (see Figure \ref{fig:cohorts}). 
For evaluating the performance of our model, DiffChest, we further partitioned 30\% of patients with manually labeled studies from the second subset to form a testing set, resulting in a total of 7,272 patients dedicated to model evaluation.

\subsection*{Image pre-processing}

Each of the radiographs used in this study was resized to $256\times256$ and zero-padded before training and testing. 
In DiffChest, each image was then normalized to the range of $[-1, 1]$.

\subsection*{Architecture}

Our architecture comprises two main components: a self-conditioned diffusion model, $p_\theta(\mathbf{x}_{t-1} | \mathbf{x}_t, \mathbf{z})$, used for both diffusion pretraining and image generation, and a feature extractor, $p_\phi(\mathbf{z}|\mathbf{x}_0)$.
To facilitate image generation, we adopted an enhanced convolutional U-Net architecture \cite{dhariwal2021diffusion}. 
This U-Net architecture can handle input resolutions of $256\times256$. 
We leveraged the efficacy of BigGAN residual blocks \cite{ronneberger2015u, brock2018large} and global attention layers \cite{wang2018non}, which were integrated into the U-Net at multiple resolution stages. 
These additions enhance the model's ability to capture intricate details and global context, resulting in improved image generation.
Adaptive group normalization layers were used in the U-Net to facilitate self-conditioning \cite{preechakul2022diffusion,dhariwal2021diffusion}.
For achieving self-conditioning, we made changes to the U-Net's normalization layers. 
Instead of using traditional normalization techniques like batch or group normalization, we adopted adaptive group normalization layers \cite{preechakul2022diffusion}.
The feature extractor $p_\phi(\mathbf{z}|\mathbf{x}_0)$ shares the same architecture as the encoder part of the previously mentioned denoising U-Net.

\subsection*{Implementation of DiffChest}

Our self-supervised model consists of a denoising U-Net and a feature encoder that we jointly train on 497,215 images. 
To prepare the data for pretraining, all extracted images are normalized and stored in a single LMDB file.
For pretraining, we used an Adam optimizer with default $\beta_1$ = 0.9, $\beta_2$ = 0.999, $\epsilon$ = 1e-8, and no weight decay to optimize the diffusion loss (Equation \ref{equ:loss}). 
The training progress was measured by the number of real images shown to DiffChest \cite{karras2017progressive}. 
We trained our model with a fixed learning rate of 1e-4 and a batch size of 12 until 200 million real radiographs were shown to the model.
Our classification head consisted of a linear classifier, i.e., a logistic regression layer, which was trained on the latent $\mathbf{z}$ space in all experiments. 
During classifier finetuning, each latent vector was normalized using a sample mean and standard deviation of the entire finetune dataset.
All computations were performed on a GPU cluster equipped with three Nvidia RTX A6000 48 GB GPUs (Nvidia, Santa Clara, Calif). 
When not otherwise specified, the code implementations were in-house developments based on Python 3.8 (\url{https://www.python.org}) and on the software modules Numpy 1.24.3, Scipy 1.10.1, and Pytorch 2.0 \cite{paszke2019pytorch}.

\subsection*{Model pretraining and finetuning}
In total, our pretraining process used a dataset of 497,215 chest radiographs from 170,717 patients.
For the subsequent classification head finetuning, we focused on a smaller set of 18,489 clinician-annotated radiographs sourced from 16,967 patients. 
We used a single logistic regression layer to classify the radiographs into 193 classes \cite{bustos2020padchest},  see Extended Data Figure \ref{fig:modelarch}b.

\subsection*{Model interpretability using visual explanations}
Visual explanations are obtained by extrapolating the latent code from $\mathbf{z}$ to $\mathbf{z}^\ast$ linearly, moving along the targeted adversarial direction, as represented by Equation \ref{equ:adv}.
The closed-form solution for Equation \ref{equ:adv} is given by:
\begin{equation}
\mathbf{z}^\ast = \mathbf{z} + \epsilon \cdot \left( \mathbf{y}^t - \sigma(\mathbf{w}^\intercal \mathbf{z} + \mathbf{b}) \right) \mathbf{w}^\intercal,
\end{equation}
where $\mathbf{w}$ and $\mathbf{b}$ are the weight and bias of the logistic regression layer, respectively, and $\sigma$ is the sigmoid function. 
The term $\mathbf{y}^t \mathbf{w}^\intercal$ increases the probability of the target class $y^t$, while the term $-\sigma(\mathbf{w}^\intercal \mathbf{z} + \mathbf{b}) \mathbf{w}^\intercal$ decreases the probability of the original class. 
To preserve the original input information, we update only the latent code $\mathbf{z}$ towards the target class, while keeping the original image attributes unchanged, i.e., 
\begin{equation}
    \mathbf{z}^\ast = \mathbf{z} + \epsilon \cdot \mathbf{y}^t \mathbf{w}^\intercal.
    \label{equ:adv_}
\end{equation}
In our experiments, we set the factor $\epsilon$ to 0.3.
The manipulated latent code $\mathbf{z}^\ast$ was then used to condition the reverse diffusion process $p_\theta(\mathbf{x}_{t-1} | \mathbf{x}_t, \mathbf{z}^\ast)$, allowing us to generate a new image $\mathbf{x}^t$. 
For this purpose, our conditional diffusion model takes inputs $(\mathbf{z}^\ast, \mathbf{x}_T)$, where $\mathbf{x}_T$ is an encoded noisy image used to initialize the diffusion process (see section \nameref{sec:input_encoding}).
To expedite sample generation, we adopt a non-Markovian DDIM sampler with 200 sampling steps proposed by Song et al. \cite{song2020denoising}.

\subsection*{Design of reader study on confounder verification}
\begin{figure}[h!]
    \centering
    \includegraphics[trim=0 50 100 0, clip, width=\textwidth]{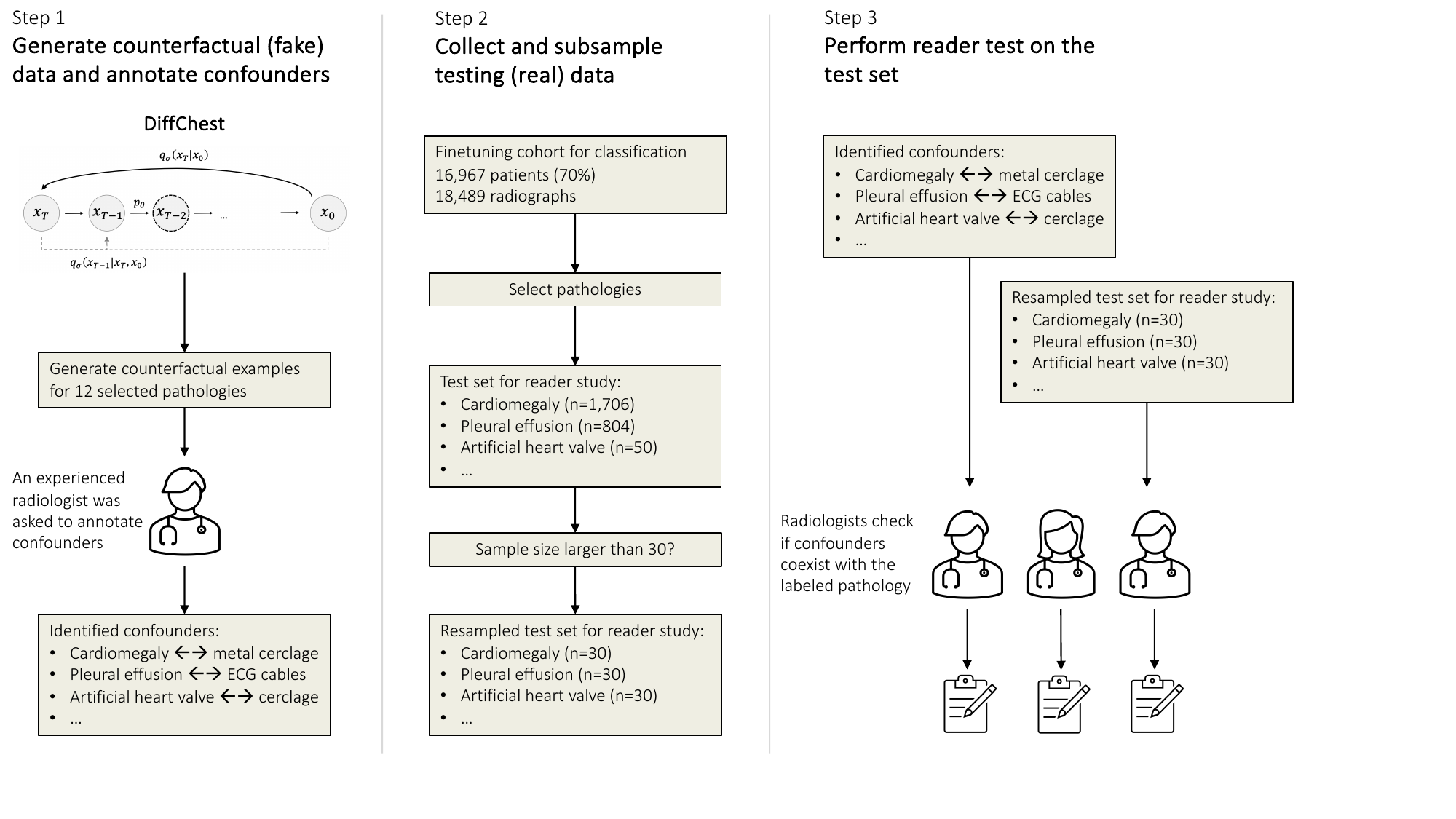}
    \captionof{table}{\textbf{Design of confounder reader study.}
    The purpose of this reader study is to validate the clinical relevance of the confounding signals visualized by our visual explanations (Figure \ref{fig:confounders}). 
    Unlike previous studies \cite{degrave2021ai}, our model is capable of discovering possible bias signals for multiple clinically relevant predictions. 
    For instance, a confident diagnosis made by an AI model on cardiopulmonary disease may not solely be based on disease features but rather due to the presence of supporting devices or sternum bindings. 
    To verify if the observed confounders indeed represent biases in the training data, we invited four experienced radiologists to review the visual explanations generated by our model (step 1), and rate the existence of the confounding signals in the training data (step 3).
    }
    \label{fig:readerstudy}
\end{figure}

Radiological confounders, as reconstructed by DiffChest, expose potential biases that are specific to the data-collecting institutions. 
To validate the accuracy and reliability of the identified confounding factors, we enlisted the expertise of four radiologists. 
These experts carried out two tasks: First, they annotated signals solely using data generated by DiffChest (as depicted in Step 1 of Extended Data Figure \ref{fig:readerstudy}); second, they verified these confounding elements using the original radiographs (as shown in Step 3 of Extended Data Figure \ref{fig:readerstudy}). 
Separately, an independent machine-learning engineer was brought in for Step 2 (refer to Extended Data Figure \ref{fig:readerstudy}) to oversee the data selection process and curate the test data set.

\subsection*{Statistics}
For each of the experiments, we calculated the ROC-AUC on the test set. 
If not otherwise stated, we extracted 95\% CI using bootstrapping with 1,000 redraws. 
The difference in ROC-AUC was defined as the $\Delta$metric.
To perform bootstrapping, we built models for the total number of N = 10,000 bootstrapping by randomly permuting predictions of two classifiers, and then computed metric differences $\Delta \text{metric}_i$ from their respective scores. 
We obtained the two-tailed p-value of individual metrics by counting all $\Delta \text{metric}_i$ above the threshold $\Delta$metric. 
Statistical significance was defined as p < 0.001.
For the clinical reader experiments, we used Fleiss' kappa to calculate inter-reader agreement between the three radiologists.

\section*{Data Availability}
All datasets used in this study are publicly available:
The chest radiography datasets from MIMIC-CXR, CheXpert, and PadChest datasets can be requested from the following URLs:
MIMIC-CXR: \url{https://physionet.org/content/mimic-cxr/2.0.0/} (requires credential access for potential users);
CheXpert: \url{https://stanfordmlgroup.github.io/competitions/chexpert/};
PadChest: \url{https://bimcv.cipf.es/bimcv-projects/padchest/}.
Testing labels and radiologists' annotations of CheXpert can be downloaded from \url{https://github.com/rajpurkarlab/cheXpert-test-set-labels}.

\section*{Code Availability}
The code and pretrained models used in this study is made fully publicly available under \url{https://github.com/peterhan91/diffchest}. 
Additionally, our demo for generating visual explanations is accessible at \url{https://colab.research.google.com/drive/1gHWCQxreE1Olo2uQiXfSFSVInmiX85Nn?usp=sharing}.

\section*{Acknowledgements}
This work is supported in part by the German Federal Ministry of Health 
(DEEP LIVER, ZMVI1-2520DAT111) and the Max-Eder-Programme of the German Cancer Aid 70113864 (received by J.N.K).

\section*{Author contributions}
T.H., J.N.K., S.N., and D.T. devised the concept of the study, D.T., L.C.H., M.S.H. and R.M.S. performed the reader tests. 
T.H. wrote the code and performed the performance studies. 
T.H. and D.T. did the statistical analysis. 
T.H., J.N.K., and D.T. wrote the first draft of the manuscript. 
All authors contributed to correcting the manuscript.

\section*{Competing interests}
J.N.K. declares consulting services for Owkin, France; DoMore Diagnostics, Norway and Panakeia, UK; furthermore he holds shares in StratifAI GmbH and has received honoraria for lectures by Bayer, Eisai, MSD, BMS, Roche, Pfizer and Fresenius.
D.T. holds shares in StratifAI GmbH. The other authors declare no competing interests.
The remaining authors declare no competing interests.

\clearpage
\bibliographystyle{naturemag}  
\bibliography{references}  

\renewcommand{\tablename}{\textbf{Extended Data Fig.}}
\setcounter{table}{0}
\clearpage

\begin{figure}[p]
    \centering
    \includegraphics[trim=0 150 30 0, clip, width=\textwidth]{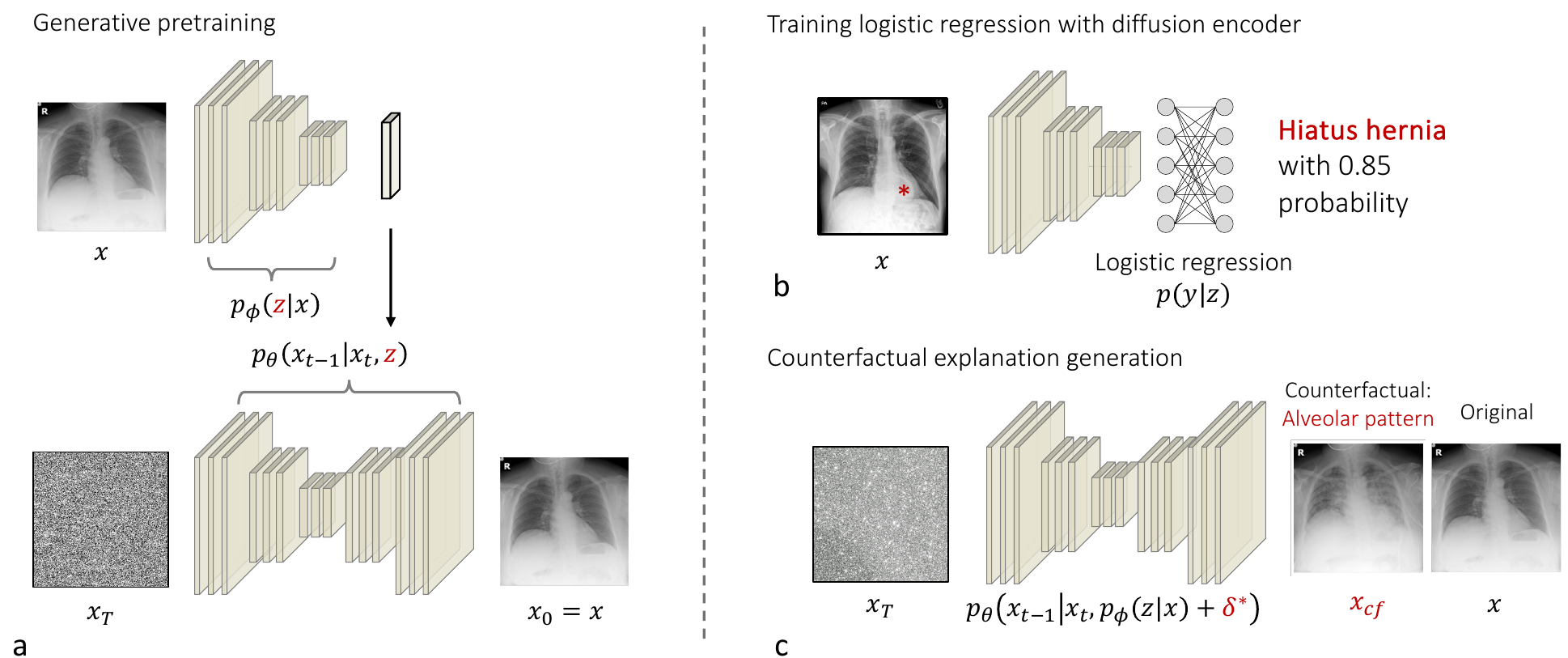}
    \captionof{table}{\textbf{Schematic of DiffChest workflows.}
    (a) DiffChest starts with self-supervised pretraining on an unlabeled dataset. 
    In contrast to current contrastive methods, DiffChest relies on the simplified diffusion objective (Equation \ref{equ:loss}) to jointly train a feature encoder and a denoising diffusion model.
    (b) It involves supervised fine-tuning on a labeled subset. 
    We only train a logistic regression classifier on top of the pretrained encoder and keep the encoder weights fixed.
    (c) Our method is both discriminative and generative. 
    We use encoder-extracted features to train a classifier and employ the diffusion model to generate visual explanations for prediction classes. 
    The latter is particularly crucial for clinicians to accept AI-assisted diagnosis, where interpretability is critical. 
    We perturb the latent feature towards a target class and utilize the diffusion model to generate a visual explanation for the target class while preserving the original information (Equation \ref{equ:adv_}).
    }
    \label{fig:modelarch}
\end{figure}

\begin{figure}[h!]
    \centering
    \includegraphics[trim=0 190 410 0, clip, width=\textwidth]{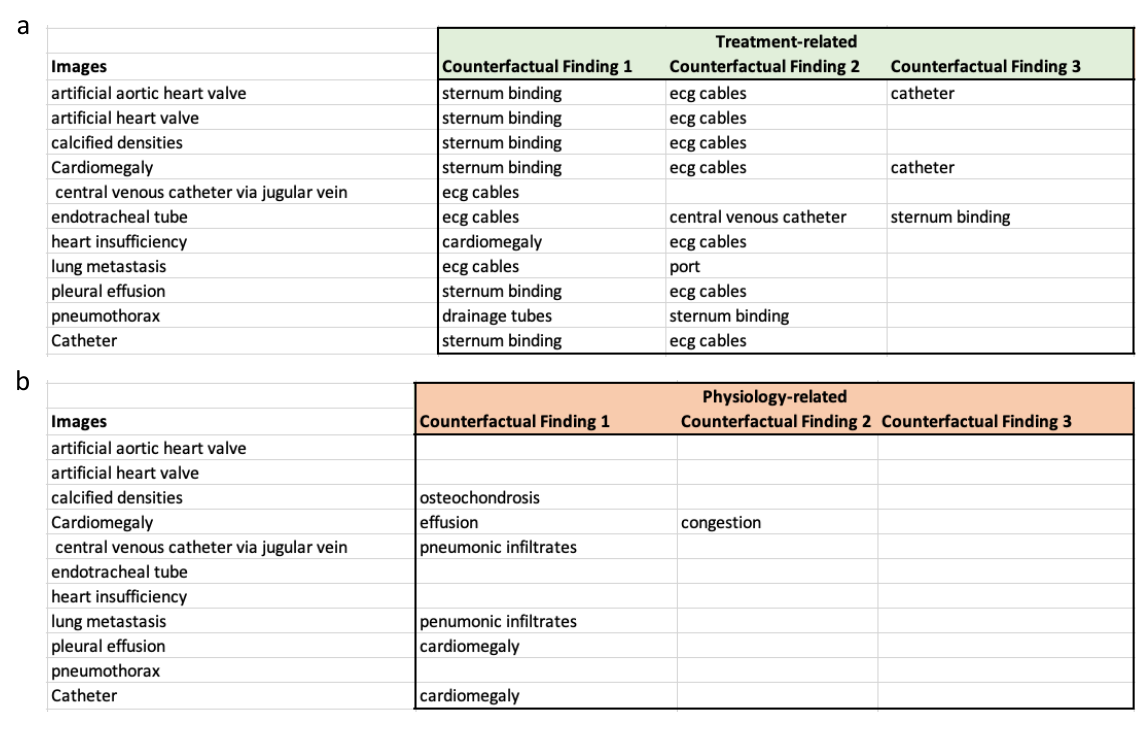}
    \captionof{table}{\textbf{Confounders identified by a radiologist using visual explanations from DiffChest.}
    This figure is the result of step-1 suggested in Extended Data Figure \ref{fig:readerstudy}.
    Both treatment \textbf{a} and physiology-related \textbf{b} confounders are identified by a radiologist using visual explanations from DiffChest.
    }
    \label{fig:daniel_con}
\end{figure}

\clearpage
\section*{Supplementary Information}
\setcounter{figure}{0}
\setcounter{table}{0}
\captionsetup[figure]{labelfont={bf},name={Supplementary Figure},labelsep=period}
\captionsetup[table]{labelfont={bf},name={Supplementary Table},labelsep=period}
\renewcommand{\thefigure}{S\arabic{figure}}
\renewcommand{\thetable}{S\arabic{table}}

\renewcommand\thesection{\Alph{section}}
\renewcommand\thesubsection{\thesection.\alph{subsection}}

\section{Extended derivations} \label{sec:ext_deri}
\subsection*{Mutual information}

Mutual information is a general way to measure dependency between two random variables.
In information theory, the mutual information can be defined as the difference of two entropies:
\begin{equation}
    I(\mathbf{z}; \mathbf{x}) = H(\mathbf{z}) - H(\mathbf{z}|\mathbf{x}) = H(\mathbf{x}) - H(\mathbf{x}|\mathbf{z}).
\end{equation}
$I(\mathbf{z}; \mathbf{x})$ is the reduction of uncertainty in $\mathbf{x}$ when $\mathbf{z}$ is observed.
In self-supervised learning, we hope to find a compressed and informative representation of the input data, i.e., maximizing the mutual information between the input and the latent representation.
However, it is hard to compute the mutual information directly due to the intractable posterior:
\begin{equation}
    \begin{aligned}
        I(\mathbf{z}; \mathbf{x}) &= H(\mathbf{z}) - H(\mathbf{z}|\mathbf{x}) \\
        &= H(\mathbf{z}) - \mathbb{E}_{(\mathbf{z, x}) \sim p(\mathbf{z, x})} \left[ -\log p(\mathbf{z} | \mathbf{x}) \right]. \\ 
    \end{aligned}
\end{equation}
In VIM \cite{chen2016infogan}, we can estimate the posterior by introducing a variational distribution $p_\phi (\mathbf{z} | \mathbf{x})$:
\begin{equation}
    \begin{aligned}
        I(\mathbf{z}; \mathbf{x}) &= H(\mathbf{z}) - \mathbb{E}_{(\mathbf{z, x}) \sim p(\mathbf{z, x})} \left[ -\log p(\mathbf{z} | \mathbf{x}) \right] \\
        &= H(\mathbf{z}) + \mathbb{E}_{(\mathbf{z, x}) \sim p(\mathbf{z, x})} \left[\log p(\mathbf{z} | \mathbf{x}) -\log p_\phi(\mathbf{z} | \mathbf{x}) + \log p_\phi(\mathbf{z} | \mathbf{x})\right] \\
        & = H(\mathbf{z}) + \mathbb{E}_{(\mathbf{z, x}) \sim p(\mathbf{z, x})} \left[D_\text{KL}\left(p(\mathbf{z} | \mathbf{x}) \, || \, p_\theta(\mathbf{z} | \mathbf{x})\right) + \log p_\phi(\mathbf{z} | \mathbf{x})\right] \\
        &\geq H(\mathbf{z}) + \mathbb{E}_{(\mathbf{z, x}) \sim p(\mathbf{z, x})} \left[ \log p_\phi(\mathbf{z} | \mathbf{x}) \right].
    \end{aligned}
    \label{equ:vlb_mi}
\end{equation}
The entropy $H(\mathbf{z})$ is typically treated as a constant, and the lower bound of mutual information is given by $\mathbb{E}_{(\mathbf{z, x}) \sim p(\mathbf{z, x})} \left[ \log p_\phi(\mathbf{z} | \mathbf{x}) \right]$ (Equation \ref{equ:mutualinfo}). 

\subsection*{Simplified diffusion objective}
Next, we will derive the simplified diffusion objective from the variational lower bound (VLB) on negative log joint likelihood $- \log p_{\theta, \phi}(\mathbf{x}_0, \mathbf{z})$ in DiffChest.
Given the forward diffusion process as $q\left(\mathbf{x}_{1:T}|\mathbf{x}_0\right)$ and our designed backward process as $p_{\theta, \phi} \left(\mathbf{x}_{1:T} |\mathbf{x}_0, \mathbf{z}\right)$, we can derive the VLB as:
\begin{equation}
    \begin{aligned}
        - \log p_{\theta, \phi}(\mathbf{x}_0, \mathbf{z}) &\leq - \log p_{\theta, \phi}(\mathbf{x}_0, \mathbf{z}) + D_{\text{KL}}\left(q\left(\mathbf{x}_{1:T}|\mathbf{x}_0\right) || \; p_{\theta, \phi} \left(\mathbf{x}_{1:T} |\mathbf{x}_0, \mathbf{z}\right)\right) \\
        &= - \log p_{\theta, \phi}(\mathbf{x}_0, \mathbf{z}) + \mathbb{E}_{\mathbf{x}_{1:T} \sim q\left(\mathbf{x}_{1:T}|\mathbf{x}_0\right)} \left[\log \frac{q\left(\mathbf{x}_{1:T}|\mathbf{x}_0\right)}{p_{\theta, \phi} \left(\mathbf{x}_{1:T}|\mathbf{x}_0, \mathbf{z}\right)} \right] \\ 
        &= \mathbb{E}_{(\mathbf{x}_{0:T}, \mathbf{z}) \sim q\left(\mathbf{x}_{0:T}, \mathbf{z}\right)} \left[\log \frac{q\left(\mathbf{x}_{1:T}|\mathbf{x}_0\right)}{p_{\theta, \phi}  \left(\mathbf{x}_{0:T}, \mathbf{z}\right)} \right] \\
        &= \mathbb{E}_q \left[ \log \frac{\prod_{t=1}^{T} q(\mathbf{x}_t | \mathbf{x}_{t-1})}{p(\mathbf{x}_T, \mathbf{z})\prod_{t=1}^{T} p_{\theta, \phi} (\mathbf{x}_{t-1} | \mathbf{x}_t, \mathbf{z})} \right] \\
        &= \mathbb{E}_q \left[ -\log p(\mathbf{x}_T, \mathbf{z}) + \sum_{t=1}^{T} \log \frac{q(\mathbf{x}_t | \mathbf{x}_{t-1})}{p_{\theta, \phi} (\mathbf{x}_{t-1}|\mathbf{x}_t, \mathbf{z})} \right] \\
        &= \mathbb{E}_q \left[ -\log p(\mathbf{x}_T, \mathbf{z}) + \sum_{t=2}^{T} \log \frac{q(\mathbf{x}_t | \mathbf{x}_{t-1})}{p_{\theta, \phi} (\mathbf{x}_{t-1}|\mathbf{x}_t, \mathbf{z})} + \log \frac{q(\mathbf{x}_1 | \mathbf{x}_0)}{p_{\theta, \phi}(\mathbf{x}_0 | \mathbf{x}_1, \mathbf{z})} \right] \\
        &= \mathbb{E}_q \left[ -\log p(\mathbf{x}_T, \mathbf{z}) + \sum_{t=2}^{T} \log \left( \frac{q(\mathbf{x}_{t-1}|\mathbf{x}_t, \mathbf{x}_0)}{p_{\theta, \phi}(\mathbf{x}_{t-1} | \mathbf{x}_t, \mathbf{z})} \cdot \frac{q(\mathbf{x}_t|\mathbf{x}_0)}{q(\mathbf{x}_{t-1} | \mathbf{x}_0)} \right) + \log \frac{q(\mathbf{x}_1 | \mathbf{x}_0)}{p_{\theta, \phi}(\mathbf{x}_0 | \mathbf{x}_1, \mathbf{z})} \right] \\
        &= \mathbb{E}_q \left[ \underbrace{\log \frac{q(\mathbf{x}_T | \mathbf{x}_0)}{p(\mathbf{x}_T, \mathbf{z})} \vphantom{\sum_{t=2}^{T}} }_{L_T} + \underbrace{\sum_{t=2}^{T} \log \frac{q(\mathbf{x}_{t-1}|\mathbf{x}_t, \mathbf{x}_0)}{p_{\theta, \phi} (\mathbf{x}_{t-1} | \mathbf{x}_t, \mathbf{z})}}_{L_t} - \underbrace{\log p_{\theta, \phi} (\mathbf{x}_0 | \mathbf{x}_1, \mathbf{z}) \vphantom{\sum_{t=2}^{T}}}_{L_0} \right]. \\
    \end{aligned}
\end{equation}
The VLB is now factorized into three terms, namely $L_0$, $L_t$, and $L_T$.
We may ignore $L_T$ since it has no learnable parameters and $\mathbf{x}_T$ is by definition a Gaussian noise.
Following, \cite{preechakul2022diffusion}, we model $L_0$ using a deterministic decoder derived from $\mathcal{N}(\mathbf{x}_0; \mu_{\theta, \phi}(\mathbf{x}_1, 1, \mathbf{z}), \mathbf{0})$. 
In DDIM's formulation, the mean of the reverse kernel $\mu_{\theta, \phi}(\mathbf{x}_t, t, \mathbf{z})$ is parameterized as
\begin{equation}
    \mu_{\theta, \phi}(\mathbf{x}_t, t, \mathbf{z}) = \frac{1}{\sqrt{\alpha_t}} \left( \mathbf{x}_t - \sqrt{1 - \alpha_t} \, \mathbf{\epsilon}_{\theta, \phi} (\mathbf{x}_t, t, \mathbf{z})\right),
    \label{equ:mean}
\end{equation}
where $\alpha_t$ is the diffusion parameter used in \cite{song2020denoising}.

Equation \ref{equ:loss} is derived from $L_t$. 
Recall, Ho et al. proved the forward kernel $q(x_{t-1} | x_t, x_0)$ follows Gaussian distributions \cite{ho2020denoising}.
The backward kernel $p_{\theta, \phi} (x_{t-1} | x_t, z)$ is also Gaussian since the step size is small. 
$L_t$ is $D_\text{KL}\left( q(x_{t-1} | x_t, x_0) \; || \; p_{\theta, \phi} (x_{t-1} | x_t, z) \right)$, since both distributions are Gaussian, we can derive the closed-form solution for $L_t$:
\begin{equation}
    \begin{aligned}
        L_t &= \mathbb{E}_{\mathbf{x}_0, \mathbf{\epsilon}} \left[ \frac{1}{2 \left\| \Sigma_{\theta, \phi} (\mathbf{x}_t, t, \mathbf{z}) \right\|_2^2} \left\| \mu_t(\mathbf{x}_t, \mathbf{x}_0) - \mu_{\theta, \phi}(\mathbf{x}_t, t, \mathbf{z}) \right\|^2  \right] \\ 
        &= \mathbb{E}_{\mathbf{x}_0, \mathbf{\epsilon}} \left[ \frac{(1 - \alpha_t)^2}{2 \alpha_t (1 - \bar{\alpha}_t) \left\| \Sigma_{\theta, \phi} \right\|_2^2} \left\| \mathbf{\epsilon}_t - \mathbf{\epsilon}_{\theta, \phi}(\mathbf{x}_t, t, \mathbf{z}) \right\|^2  \right], 
    \end{aligned}
    \label{equ:full_loss}
\end{equation}
where $\Sigma_{\theta, \phi}(\mathbf{x}_t, t, \mathbf{z})$ is the variance and $\mathbf{\epsilon}_t$ is the Gaussian noise added at step $t$.
We can get the simplified diffusion objective (Equation \ref{equ:loss}) by neglecting the weighting term in Equation \ref{equ:full_loss}.

\subsection*{Mutual information maximized pretraining}
We have proven in the previous section that optimizing our diffusion pretraining loss, i.e., Equation \ref{equ:loss}, is equivalent to maximizing the VLB of the likelihood $\log p_{\theta, \phi}(\mathbf{x}_0, \mathbf{z}) = \log p_{\theta}(\mathbf{x}_0) + \log p_{\theta, \phi}(\mathbf{z} | \mathbf{x}_0)$.
The first term, $\log p_{\theta}(\mathbf{x}_0)$, represents the log likelihood of the input data, while the second term, $\log p_{\theta, \phi}(\mathbf{z} | \mathbf{x}_0)$, corresponds to the VLB of the mutual information between the input and the latent representation (Equation \ref{equ:vlb_mi}).
Therefore, optimizing our diffusion pretraining loss is equivalent to maximizing both the data likelihood and the mutual information between the input $\mathbf{x}$ and its latent representation $\mathbf{z}$.

\section{Visual explanation generation}
\subsection*{Input encoding} \label{sec:input_encoding}
\begin{figure}[h]
    \centering
    \includegraphics[trim= 0 70 245 0, clip, width=\textwidth]{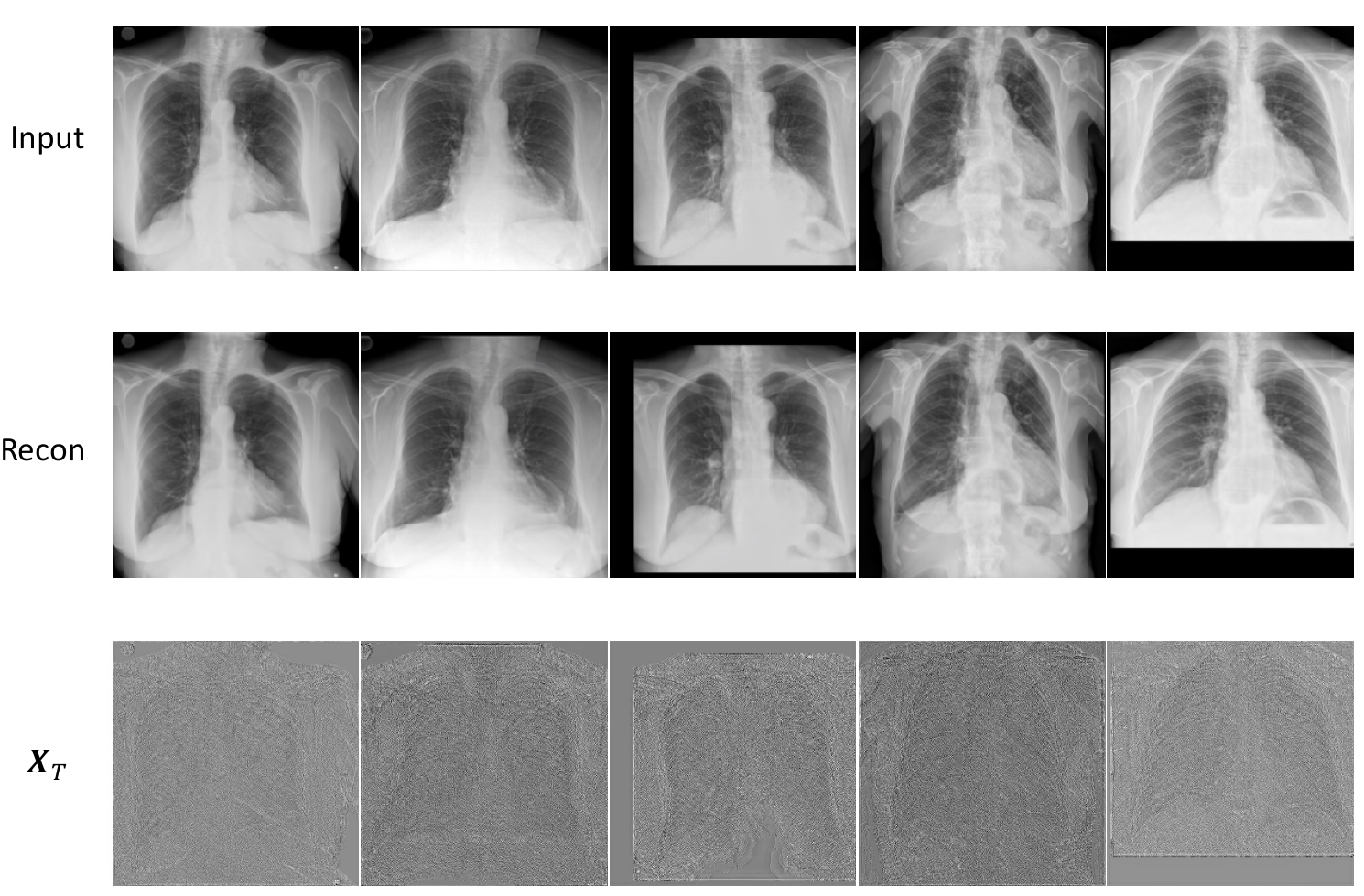}
    \caption{\textbf{Encoding input images to diffusion latent space.}
    We embed the input images into the diffusion latent space following a deterministic multistep process (Equation \ref{equ:ddim_enc}). 
    The top row shows the original images, and the bottom row shows the corresponding latent codes $\mathbf{x}_T$. 
    In the middle, we visualize reconstructions of the original images from the latent codes.
    }
    \label{fig:recon}
\end{figure}

The latent space of DiffChest consists of two parts: a 512-dimensional latent code $\mathbf{z}$ and a 256$\times$256 noise map $\mathbf{x}_T$, as depicted in Extended Data Figure \ref{fig:modelarch}.
To generate visual explanations, we first need to encode the input image $\mathbf{x}_0$ into both latent spaces. Embedding $\mathbf{x}_0$ into $\mathbf{z}$ space is accomplished using our trained feature extractor $p_\phi(\mathbf{z}|\mathbf{x}_0)$. 
Similarly, we encode $\mathbf{x}_0$ into $\mathbf{x}_T$ space by running the generative process in our diffusion model backward. 
This process is deterministic as we utilized a DDIM sampler for sample generation:
\begin{equation}
    \mathbf{x}_{t} = \sqrt{\alpha_{t}} \, \mu_{\theta, \phi}(\mathbf{x_{t-1}}, t-1, \mathbf{z}) + \sqrt{1 - \alpha_t} \, \mathbf{\epsilon}_{\theta, \phi} ( \mathbf{x}_{t-1}, t-1, \mathbf{z}).
    \label{equ:ddim_enc}
\end{equation}
Here, the mean $\mu_{\theta, \phi}(\mathbf{x}_t, t, \mathbf{z})$ is defined in Equation \ref{equ:mean}. 
In our experiments, we set the number of encoding steps to 250.
After encoding, we can use both the encoded $\mathbf{z}$ and $\mathbf{x}_T$ (as shown in Figure \ref{fig:recon}) to reconstruct the input image. 
As depicted in Figure \ref{fig:recon}, sharp and high-fidelity reconstructions can be achieved by using our DiffChest with just 100 sampling steps.
 
\section{DiffChest is data efficient} \label{sec:data_efficiency}

\begin{figure}[h!]
    \centering
    \scalebox{0.7}{
    \includegraphics[trim=0 100 630 0, clip, width=\textwidth]{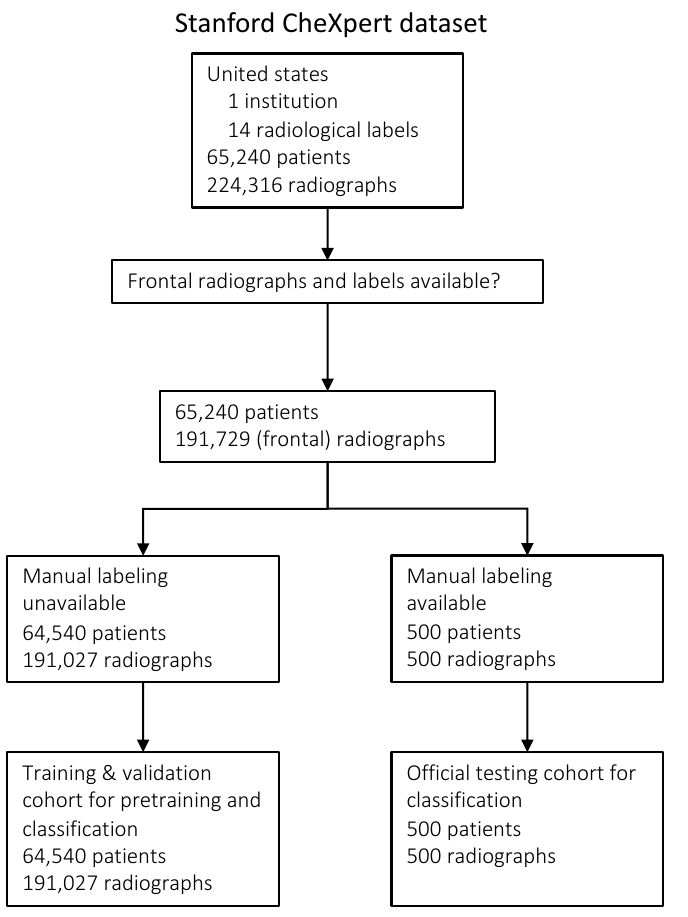}}
    \caption{\textbf{Data flow chart for CheXpert dataset.}
    }
    \label{fig:chexpert_data}
\end{figure}

\begin{figure}[h!]
    \centering
    \includegraphics[trim=0 100 100 0, clip, width=\textwidth]{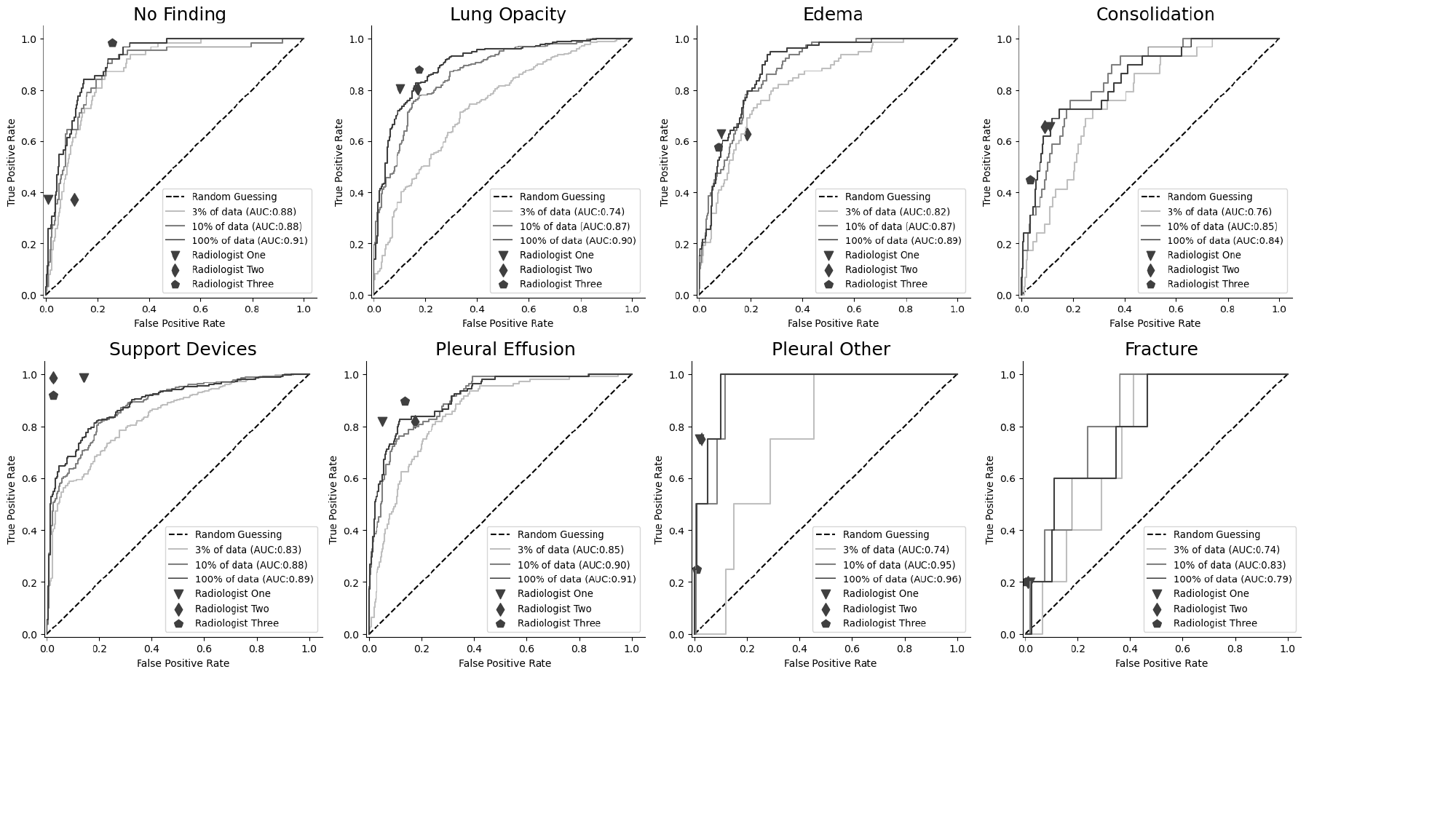}
    \caption{\textbf{DiffChest is data efficient.}
    After completing the diffusion pretraining, we compared three models: classification heads finetuned using 100\% ($N=7,943$), 10\% ($N=794$), and 3\% ($N=238$) of the CheXpert training data. 
    To evaluate and compare the performance, we used receiver operating characteristic (ROC) curves for these models against benchmarking radiologists provided by the CheXpert competition. 
    The testing cohort consisted of $N=500$ radiographs.
    }
    \label{fig:dataefficiency}
\end{figure}

Learning from few annotated datasets poses a significant challenge in medical AI, as the predictive performance of models generally improves with larger training datasets \cite{lu2021data} and as high-quality medical data for training purposes is scarce. 
To address this issue, we investigated the potential of diffusion pretraining to mitigate performance losses when using only a small subset of radiographs from the CheXpert dataset for model tuning (the data flow is given in Figure \ref{fig:chexpert_data}). 
We employed CheXpert for our study due to the availability of high-quality testing labels and annotations from three board-certified radiologists \cite{irvin2019chexpert}.
Our findings demonstrate that DiffChest's performance remains stable and almost on par with radiologists when finetuning the model using only 10\% of the total data (Figure \ref{fig:dataefficiency}). 
However, it is worth noting that all radiologists clearly outperformed DiffChest on the classification of supporting devices. 
We attribute this to the presence of label noise within the CheXpert dataset, as DiffChest achieved an almost perfect AUC when classifying device labels such as single chamber device, dual chamber device, and pacemaker in the PadChest dataset (Figure \ref{fig:vschexzero}).
Even when downsampled to only 3\% of the total data, DiffChest maintains high AUC scores of at least 0.800 for No Finding, Edema, Supporting Devices, and Pleural Effusion.

\clearpage
\section{Supplementary tables and figures}
\begin{longtable}{lllr}
    \toprule
                                        Patient subgroup &            Diffchest (95\% CI) &             CheXzero (95\% CI) &  p-value \\
    \midrule
                                     sternotomy & 0.988 (0.982, 0.994) & 0.810 (0.772, 0.847) &    <1e-4\\
                              endotracheal tube & 0.978 (0.968, 0.986) & 0.987 (0.983, 0.990) &    0.820 \\
                                      pacemaker & 0.978 (0.967, 0.989) & 0.992 (0.989, 0.995) &    0.660 \\
                                       nsg tube & 0.977 (0.970, 0.983) & 0.928 (0.909, 0.949) &    0.147 \\
                            dual chamber device & 0.977 (0.968, 0.985) & 0.974 (0.967, 0.979) &    0.960 \\
       central venous catheter via jugular vein & 0.964 (0.944, 0.982) & 0.975 (0.966, 0.984) &    0.823 \\
                          single chamber device & 0.946 (0.927, 0.967) & 0.973 (0.966, 0.979) &    0.684 \\
                             mammary prosthesis & 0.944 (0.903, 0.987) & 0.532 (0.444, 0.630) &    <1e-4\\
                               pleural effusion & 0.943 (0.933, 0.954) & 0.971 (0.964, 0.978) &    0.195 \\
                              tracheostomy tube & 0.940 (0.894, 0.983) & 0.960 (0.933, 0.986) &    0.724 \\
                            heart insufficiency & 0.936 (0.922, 0.951) & 0.936 (0.920, 0.951) &    0.992 \\
                                   cardiomegaly & 0.899 (0.890, 0.909) & 0.903 (0.894, 0.911) &    0.795 \\
                               alveolar pattern & 0.894 (0.874, 0.913) & 0.898 (0.877, 0.918) &    0.862 \\
                                  hiatal hernia & 0.893 (0.864, 0.924) & 0.826 (0.792, 0.861) &    0.083 \\
                               hilar congestion & 0.885 (0.855, 0.914) & 0.903 (0.882, 0.925) &    0.651 \\
                                         goiter & 0.874 (0.830, 0.920) & 0.605 (0.542, 0.672) &    <1e-4\\
                                   gynecomastia & 0.872 (0.842, 0.904) & 0.563 (0.508, 0.614) &    <1e-4\\
    central venous catheter via subclavian vein & 0.872 (0.824, 0.919) & 0.917 (0.889, 0.946) &    0.414 \\
                                       kyphosis & 0.869 (0.844, 0.891) & 0.722 (0.690, 0.753) &    <1e-4\\
                              aortic elongation & 0.865 (0.852, 0.878) & 0.675 (0.658, 0.693) &    <1e-4\\
                           interstitial pattern & 0.850 (0.830, 0.868) & 0.843 (0.822, 0.863) &    0.739 \\
                        osteosynthesis material & 0.840 (0.770, 0.917) & 0.576 (0.495, 0.652) &    <1e-4\\
                                      emphysema & 0.829 (0.786, 0.873) & 0.801 (0.749, 0.854) &    0.573 \\
                               suboptimal study & 0.827 (0.795, 0.858) & 0.783 (0.747, 0.822) &    0.197 \\
                                         normal & 0.825 (0.816, 0.834) & 0.752 (0.741, 0.763) &    <1e-4\\
                      apical pleural thickening & 0.823 (0.802, 0.844) & 0.641 (0.611, 0.673) &    <1e-4\\
                   descendent aortic elongation & 0.819 (0.757, 0.882) & 0.789 (0.745, 0.838) &    0.647 \\
                            hypoexpansion basal & 0.815 (0.737, 0.892) & 0.806 (0.753, 0.859) &    0.916 \\
                                     copd signs & 0.814 (0.800, 0.828) & 0.755 (0.738, 0.772) &    <1e-4\\
                    costophrenic angle blunting & 0.808 (0.785, 0.831) & 0.798 (0.775, 0.823) &    0.636 \\
                           aortic atheromatosis & 0.805 (0.784, 0.824) & 0.699 (0.675, 0.723) &    <1e-4\\
                                         bullas & 0.805 (0.729, 0.878) & 0.557 (0.464, 0.653) &    <1e-4\\
                                    volume loss & 0.804 (0.769, 0.840) & 0.730 (0.692, 0.770) &    0.027 \\
                                     osteopenia & 0.798 (0.725, 0.869) & 0.673 (0.602, 0.749) &    0.051 \\
                                    atelectasis & 0.792 (0.753, 0.831) & 0.814 (0.781, 0.851) &    0.507 \\
                                 tracheal shift & 0.791 (0.700, 0.871) & 0.699 (0.630, 0.765) &    0.137 \\
                             pleural thickening & 0.790 (0.715, 0.867) & 0.822 (0.768, 0.878) &    0.601 \\
                                chronic changes & 0.789 (0.775, 0.804) & 0.654 (0.636, 0.671) &    <1e-4\\
                 vertebral degenerative changes & 0.779 (0.759, 0.800) & 0.644 (0.616, 0.672) &    <1e-4\\
                                    azygos lobe & 0.779 (0.706, 0.852) & 0.465 (0.352, 0.579) &    <1e-4\\
                          tuberculosis sequelae & 0.778 (0.696, 0.857) & 0.642 (0.557, 0.725) &    0.032 \\
                      diaphragmatic eventration & 0.776 (0.730, 0.825) & 0.602 (0.547, 0.652) &    <1e-4\\
                                      pneumonia & 0.770 (0.745, 0.796) & 0.817 (0.793, 0.841) &    0.038 \\
                                   air trapping & 0.765 (0.740, 0.791) & 0.590 (0.561, 0.618) &    <1e-4\\
                       bronchovascular markings & 0.764 (0.713, 0.818) & 0.686 (0.640, 0.730) &    0.038 \\
                        hemidiaphragm elevation & 0.761 (0.724, 0.797) & 0.847 (0.817, 0.877) &    0.010 \\
                                      scoliosis & 0.753 (0.731, 0.774) & 0.646 (0.623, 0.672) &    <1e-4\\
                                 bronchiectasis & 0.752 (0.708, 0.795) & 0.751 (0.710, 0.794) &    0.966 \\
                 vertebral anterior compression & 0.745 (0.701, 0.791) & 0.636 (0.586, 0.686) &    0.006 \\
                                suture material & 0.745 (0.690, 0.800) & 0.558 (0.503, 0.613) &    <1e-4\\
                            callus rib fracture & 0.743 (0.706, 0.781) & 0.659 (0.621, 0.697) &    0.008 \\
                                 pulmonary mass & 0.740 (0.658, 0.818) & 0.904 (0.864, 0.945) &    0.008 \\
                                  consolidation & 0.734 (0.659, 0.806) & 0.888 (0.846, 0.928) &    0.004 \\
                                    infiltrates & 0.720 (0.689, 0.752) & 0.772 (0.747, 0.800) &    0.044 \\
                            laminar atelectasis & 0.717 (0.684, 0.751) & 0.689 (0.657, 0.720) &    0.291 \\
                                  nipple shadow & 0.715 (0.658, 0.771) & 0.633 (0.574, 0.688) &    0.077 \\
                     vascular hilar enlargement & 0.714 (0.685, 0.745) & 0.631 (0.598, 0.665) &    0.002 \\
                                  fibrotic band & 0.712 (0.682, 0.748) & 0.633 (0.601, 0.665) &    0.003 \\
                      aortic button enlargement & 0.704 (0.631, 0.781) & 0.700 (0.639, 0.762) &    0.962 \\
                              lobar atelectasis & 0.690 (0.596, 0.781) & 0.807 (0.734, 0.879) &    0.099 \\
                              hilar enlargement & 0.686 (0.634, 0.738) & 0.679 (0.623, 0.735) &    0.852 \\
                                          metal & 0.676 (0.598, 0.751) & 0.536 (0.462, 0.610) &    0.007 \\
                             hyperinflated lung & 0.673 (0.589, 0.760) & 0.748 (0.664, 0.829) &    0.257 \\
                            flattened diaphragm & 0.662 (0.583, 0.746) & 0.567 (0.476, 0.660) &    0.172 \\
                                   rib fracture & 0.650 (0.541, 0.752) & 0.771 (0.689, 0.855) &    0.093 \\
                                      unchanged & 0.648 (0.629, 0.667) & 0.544 (0.524, 0.566) &    <1e-4\\
                                   pseudonodule & 0.635 (0.591, 0.676) & 0.510 (0.468, 0.549) &    <1e-4\\
                            calcified granuloma & 0.632 (0.589, 0.678) & 0.471 (0.423, 0.518) &    <1e-4\\
                                      granuloma & 0.619 (0.535, 0.700) & 0.476 (0.382, 0.563) &    0.040 \\
                            calcified densities & 0.618 (0.555, 0.683) & 0.528 (0.471, 0.593) &    0.065 \\
                           ground glass pattern & 0.598 (0.494, 0.717) & 0.655 (0.553, 0.760) &    0.427 \\
                                         nodule & 0.587 (0.541, 0.635) & 0.580 (0.531, 0.630) &    0.854 \\
                              increased density & 0.567 (0.525, 0.608) & 0.680 (0.638, 0.719) &    <1e-4\\
    \bottomrule
    \caption{Performance comparison between DiffChest and CheXzero.}
    \label{tab:p_ci}
\end{longtable}

\clearpage

\begin{figure}[h!]
    \centering
    \includegraphics[trim=0 0 0 0, clip, width=\textwidth]{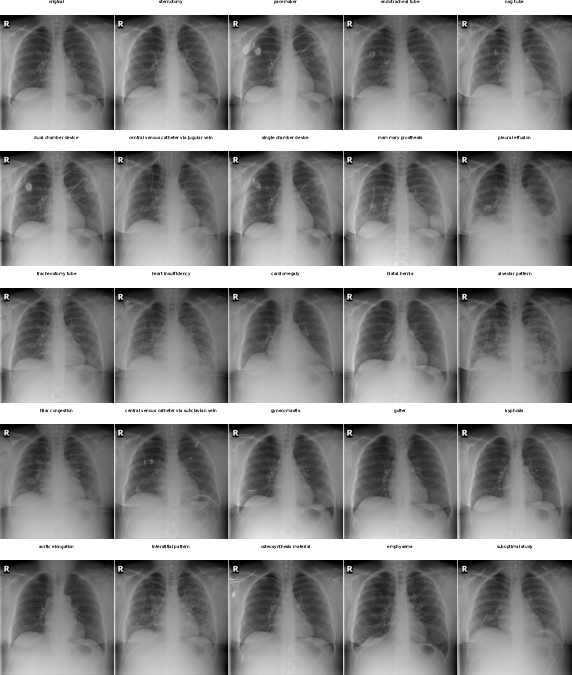}
    \caption{\textbf{Visual explanations from DiffChest for the 1st to the 24th clinical finding predictions shown in Figure \ref{fig:vschexzero}.}
    We visualize the original image in the top left panel of this image grid. 
    }
    \label{fig:more_counterfactuals_24}
\end{figure}

\begin{figure}[h!]
    \centering
    \includegraphics[trim=0 0 0 0, clip, width=\textwidth]{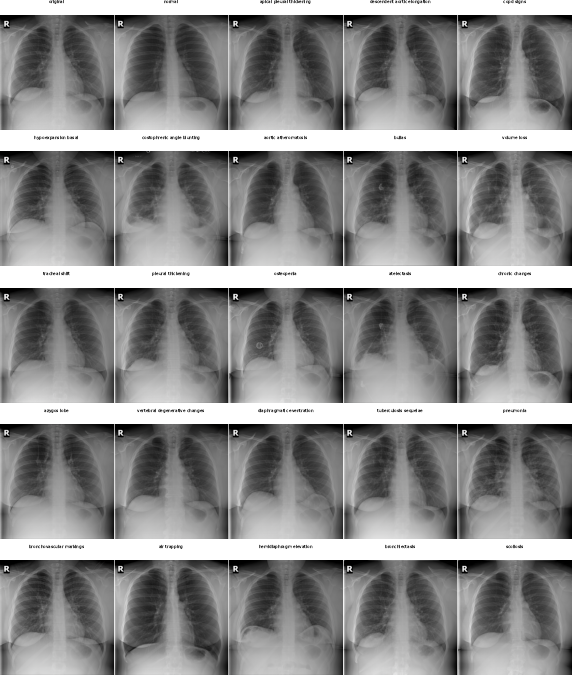}
    \caption{\textbf{Visual explanations from DiffChest for the 25th to the 48th clinical finding predictions shown in Figure \ref{fig:vschexzero}.}
    We visualize the original image in the top left panel of this image grid. 
    }
    \label{fig:more_counterfactuals_48}
\end{figure}

\begin{figure}[h!]
    \centering
    \includegraphics[trim=0 0 0 0, clip, width=\textwidth]{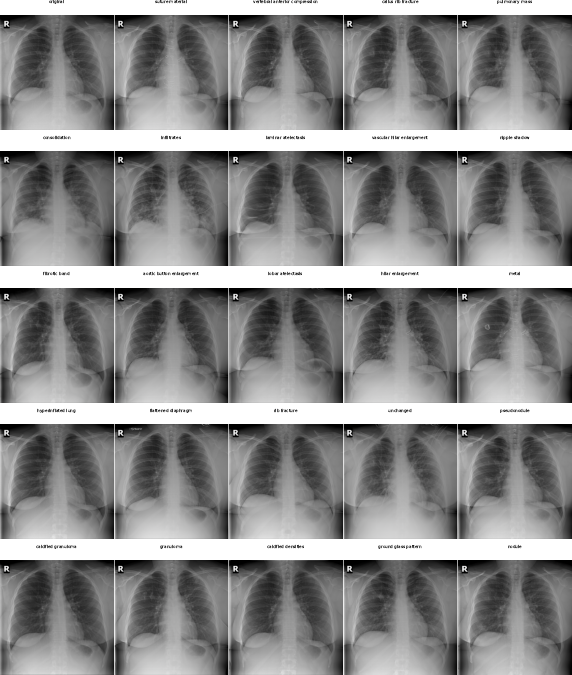}
    \caption{\textbf{Visual explanations from DiffChest for the 49th to the 72th clinical finding predictions shown in Figure \ref{fig:vschexzero}.}
    We visualize the original image in the top left panel of this image grid. 
    }
    \label{fig:more_counterfactuals_72}
\end{figure}

\end{document}